\documentclass[journal]{IEEEtran}
\usepackage{amsmath,amsfonts}
\usepackage{array}
\usepackage[caption=false,font=normalsize,labelfont=sf,textfont=sf]{subfig}
\usepackage{textcomp}
\usepackage{stfloats}
\usepackage{url}
\usepackage{graphicx}
\usepackage{cite}
\usepackage{color}
\usepackage{diagbox}
\usepackage[ruled,linesnumbered]{algorithm2e}

\begin{document}
\title{Dual-Arm Robot-Assisted Dressing with Differentiable Clothing Simulation}
\author{Yiming Liu, Lijun Han, and Hesheng Wang, \emph{Senior Member, IEEE}% <-this % stops a space
\thanks{*This work was supported in part by the Natural Science Foundation of China under Grant 62225309, 62073222, U21A20480 and 62361166632, in part by the Open Research Projects of Zhejiang Lab under Grant 2022NB0AB01.}% <-this % stops a space
\thanks{Y. Liu, L. Han, and H. Wang are with the Department of Automation, Shanghai Jiao Tong University, Shanghai 200240, China (e-mail: lym5634@sjtu.edu.cn; lijun\_han@sjtu.edu.cn; wanghesheng@sjtu.edu.cn).}%
}

\markboth{Journal of \LaTeX\ Class Files,~Vol.~14, No.~8, August~2021}%
{Shell \MakeLowercase{\textit{et al.}}: A Sample Article Using IEEEtran.cls for IEEE Journals}
\IEEEpubid{0000--0000/00\$00.00~\copyright~2021 IEEE}
\maketitle

\begin{abstract}
The development of assistive robots for dressing tasks serves to augment human convenience and improve the quality of life for individuals with physical impairments.  
However, due to the intricate contact interactions between garments and the human limbs during dressing, most robot-assisted dressing algorithms can only achieve partial dressing, struggling to handle with the remaining garments under contact constraints.
To overcome this challenge, we propose a novel robotic dressing control algorithm that integrates real-time differentiable clothing simulation. 
The simulation algorithm employs explicit iterative scheme with intentionally introduced high-order bias to enhance computational efficiency while maintaining stability under large time-step conditions.  
Through simulation, we resolve the garment state under contact constraints, which then enables a multi-stage control strategy for successful dressing assistance.  
To further improve real-time performance, a constrained local model is introduced along with its corresponding optimization solver, permitting high-frequency local compensation for the differentiable simulation based global controller.  
By comparing with the Projective Dynamics algorithm in simulation, we validated the real-time computational advantages of our differentiable simulation. Physical dressing experiments were conducted on users with various poses and motion patterns, conclusively demonstrating the feasibility and efficacy.  

\end{abstract}

\section{Introduction}
With the development of artificial intelligence, assistive robots \cite{NanavatiPhysicallyAssistiveRobots2023, LiuFromScreensTo2025} are increasingly used in daily life, which especially enhances the quality of life for elderly and mobility-impaired individuals. Among different daily tasks, robot-assisted dressing is one of the most common yet critical tasks. As the dressing process necessitates the manipulation of deformable garments while maintaining indirect physical interaction with humans, this task encompasses challenges in both deformable object manipulation and human-robot collaboration, rendering it a representative and demanding benchmark for cooperative robotics. Advancements in this domain hold substantial implications for improving assistive robotic capabilities.
\par
During dressing process, complex contact interactions arise between garments and the human limbs, especially when partial structures of the clothing have been worn completely. For garments like coats, the donned sleeve establishes tight physical contact \cite{EricksonWhatDoesPerson2017}, providing support and constraining the overall motion and deformation of the garment. Approaches \cite{EricksonDeepHapticModel2018, KapustaPersonalizedCollaborativePlans2019, ZhuDoYouNeed2024} have overlooked these constraints by simplifying garments as independent armholes, thereby failing to address the secondary sleeve dressing problem. To properly account for deformable garment constraints, it is essential to formulate an appropriate garment model, leveraging clothing simulation to predict contact deformation in constrained dressing tasks.
\par
The effectiveness of assisted dressing is influenced by the user's pose and motion patterns. Previous work \cite{GaoIterativePathOptimisation2016, QieCrossDomainRepresentation2023, YamasakiPersonalizedAssistAs2024} treats assisted dressing as a synchronized dual-arm trajectory tracking problem. However, when the user's arms are naturally extended beyond the garment's shoulder width, the garment is unable to be directly draped over both arms, causing these methods to converge to infeasible local minima and ultimately fail. Moreover, during the garment donning process, human arms may exhibit either passive compliance to follow the garment or active cooperation with robots, which dynamically alter the dressing trajectory and consequently affect overall dressing performance. To accommodate varying user's state, a coordinated dual-arm control strategy is valuable to guide the garment into the proper position, enabling sequential completion of the dressing task of two sleeves.
\par
In this paper, we propose a real-time differentiable clothing simulation algorithm and a corresponding multi-stage goal-guided control strategy to address the complex human-garment interaction and diverse user poses, culminating in a comprehensive dual-arm dressing assistance system. 
To effectively utilize clothing contact simulation, we develop a novel explicit iterative algorithm with dry friction, enabling rapid differentiable clothing simulation for real-time control applications. 
By incorporating predicted human motion trajectories, the multi-stage control strategy facilitates dual-sleeve donning while accommodating garment constraints across diverse human motion patterns. 
Through simplified local modeling, an optimal controller with local compensation is devised to enhance high-frequency dressing assistance. 
Experimental results validate the proposed methodology, demonstrating successful dressing assistance across different coats without prior models.
\par
The contributions are summarized as follows:
\IEEEpubidadjcol
\begin{itemize}
    \item A novel explicit differentiable clothing simulation algorithm with dry friction is designed for real-time simulation and optimal control. By decoupling the position and velocity and incorporating distinct high-order bias, we achieve fast-converging explicit iteration that rapidly attains equilibrium states  with large time steps.
    \item We propose a multi-stage objective-driven strategy to guide each sleeve donned by the arm separately for coats. By integrating human motion prediction, this framework dynamically adapts to diverse human intents while ensuring coordinated dual-arm manipulation.
    \item To improve the real-time performance of the simulation-based controller, we introduce a localized solver for high-frequency corrective adjustments. A linearized dressing model with state feasibility constraints, coupled with an active-set optimization solver is developed to achieve rapid response to unpredictable human motions.
\end{itemize}

\section{Related work}
\subsection{Robot-Assisted Dressing} 
As a fundamental task, robot-assisted dressing has received significant attention in recent years. To dress an upper-body garment, the conventional approaches utilize the arm poses and directly generate the robot trajectories through rules \cite{ZhangProbabilisticRealTime2019, KuribayashiMotionGenerationFor2024} or learning \cite{PignatLearningAdaptiveDressing2017, JoshiAFrameworkFor2019}. These methods struggle with fabric constraints and are typically limited to donning the first sleeve.
Several researches \cite{KapustaPersonalizedCollaborativePlans2019, CleggLearningToCollaborateS2020, ZhangLearningGarmentManipulation2022} have explored hospital gown dressing and achieved notable success. Yet, gowns present a simplified scenario as their loose fit and rear opening allow for straightforward front-side donning. 
With the development of simulation, Sun et al. \cite{SunForceConstrainedVisual2024} proposed a vision-based reinforcement learning approach to learn the dress strategy offline. Due to the lack of real-time simulation, this method loses the accurate clothing state and is ineffective for varying constraints induced by garment deformation and human motions. 
For conventional coats, it remains challenging to dress two sleeves simultaneously, particularly when the initial arm span exceeds shoulder width.
\par

\subsection{Cloth Simulation}
In physics-based cloth simulation, garments are typically modeled as nodes with local constraints, for which the implicit Euler integration is employed to achieve robust large-step simulation \cite{BaraffLargeStepsIn1998}. This formulation can be framed as an energy minimization problem \cite{LiuFastSimulationMass2013}, ensuring system convergence and iterative stability.
To enhance computational efficiency, Projective Dynamics (PD) \cite{BouazizProjectiveDynamicsFusing2023,NarainADMMProjectiveDynamics2016,LiuQuasiNewtonMethods2017} decompose the iterations into global linear systems and local nonlinear constraints, utilizing the global/local solver alternately to address the nonlinearity of implicit integration.
Another popular approach, Position-Based Dynamics (PBD) \cite{MullerPositionBasedDynamics2007, BenderPositionBasedSimulation2014, MacklinXPBDPositionbased2016}, projects constraints into a Gauss-Seidel like fashion, enabling highly parallelized constraint resolution at the positional level.
Despite progress, the computational cost of implicit iterations hinders real-time applications. 
Previous work \cite{BaraffLargeStepsIn1998} attempted to leverage gradient to estimate the results of implicit integration, circumventing the nonlinear equation solving. However, this idea fails to adequately handle static friction and limits subsequent development. More recently, Kotsovolis et al. \cite{KotsovolisGarmentDiffusionModels2025} employed diffusion models to learn garment dynamics for dressing assistance. Nevertheless, the complex human-robot interactions inherent in the dressing process pose significant challenges for real-world data acquisition.
\par

\subsection{Differentiable Cloth Simulation} 
The incorporation of gradient information within simulation frameworks has demonstrated significant potential in parameter identification \cite{LiDiffClothDifferentiableCloth2023}, motion planning \cite{QiaoDifferentiableSimulationOf2021}, and neural network training \cite{LiangDifferentiableClothSimulation2019}. This versatility has  led to the growing adoption of differentiable cloth simulation techniques in recent research.
In the domain of PD, Du et al. \cite{DuDiffPDDifferentiableProjective2022} developed the DiffPD, a differentiable simulator that formulates derivatives through backward iterations analogous to the  PD solvers. Building upon this foundation, DiffCloth \cite{LiDiffClothDifferentiableCloth2023} extended these capabilities to incorporate nodal collision and frictional contact \cite{MichaelProjectiveDynamicsWith2020}, including the computation of dry friction gradients. To solve cloth self-collision and interpenetration, DiffClothAI \cite{YuDiffClothAIDifferentiableCloth2023} derived the differentiable formulation of Incremental Potential Contact (IPC) \cite{LiIncrementalPotentialContact2020}, expanding the applicability of differentiable cloth simulation.
In this work, we extend our cloth simulation method from DiffCloth to include the computation of gradients, which enhances its garment control capabilities for dressing.

\section{Architecture and Problem Formulation}
\subsection{Problem Formulation}
In this work, we study the robot-assisted dressing problem for coats. As shown in Fig. \ref{pipeline}, two manipulators are controlled to pick the both sides of the coat's collar. The user to be dressed stands or sits in front of the coat and faces to the same direction with two arms opening naturally. Two RGBD cameras are placed on the front and rear sides of the platform to observe the state of the user and the coat. The dressing task aims to put the sleeves over the corresponding arms and pull the clothing armholes to the corresponding shoulders.
\par
To model the problem in mathematical form, we construct the tetrahedral grids with $n$ nodes from point cloud to describe the garment state. The $i-th$ nodal position of the grids is denoted as $\boldsymbol{p}_i \in \mathbb{R}^3, i=1, \cdots, n$. For each sleeve, the center and the orientation vector of the armhole are employed to represent the sleeve state, which is written as $\boldsymbol{x}_{ci}=(\boldsymbol{p}_{ci}, \boldsymbol{v}_{ci}),  i\in\{l,r\}$. Denote the positions of two end-effectors as $\boldsymbol{p}_{rl}, \boldsymbol{p}_{rr}$. Nodes with distances less than $l_p$ are seen as the picked nodes and move along with the actuators through spring constraints. For arms of the user, we extract the positions of shoulders, elbows, and hands as the arm states, which are denoted as $\boldsymbol{p}_{si}, \boldsymbol{p}_{ei}, \boldsymbol{p}_{hi}, i\in\{l,r\}$. 
\par
Referred to \cite{ZhuDoYouNeed2024}, we define the dressing coordinates around two arms to calculate the task progress for two sleeve states. The dressing coordinate $(s_i, l_i, \theta_i), i\in\{l,r\}$ is a cylinder coordinate, composed of the progress scalar, sleeve-to-arm distance, and rotation angle. The progress scalar $s_i$ is a scalar quantity to indicate the progress from  $\boldsymbol{p}_{hi}$ to $\boldsymbol{p}_{si}$, with $s_i=0$ at the hand, $s_i=1$ at the elbow, and $s_i=2$ at the shoulder. The sleeve not covered to the arm is defined as $s_i=0$. The axis of the cylinder coordinate  system is consisted of the axes of the forearm and upper-arm, which are connected by a tangent arc to keep smooth. $l_i$ is calculated as the distance to the axis while $\theta_i$ is the corresponding angle, with the reference $0^\circ$ directed to $(\boldsymbol{p}_{ei} - \boldsymbol{p}_{hi})\times(\boldsymbol{p}_{si} - \boldsymbol{p}_{ei})$.
\par
Thus, we define the following control problem:
\par
\textit{Problem}: Given the observation of two RGBD cameras, the body size of the user, and the material property of the coat (including mass, stiffness, and friction coefficient with human), extract and track the coat grids and the arm states as the input. With the sleeve state calculated from the grids, design a controller to control the position of two end-effectors $\boldsymbol{p}_{rl}, \boldsymbol{p}_{rr}$ to drive the dressing coordinates to $s_i=2, i\in\{l, r\}$ for two sleeves.
\subsection{Overall View of Proposed Framework}
\begin{figure*}[htbp]\centering 
        \includegraphics[width=0.95\textwidth,trim=1 1 1 1,clip]{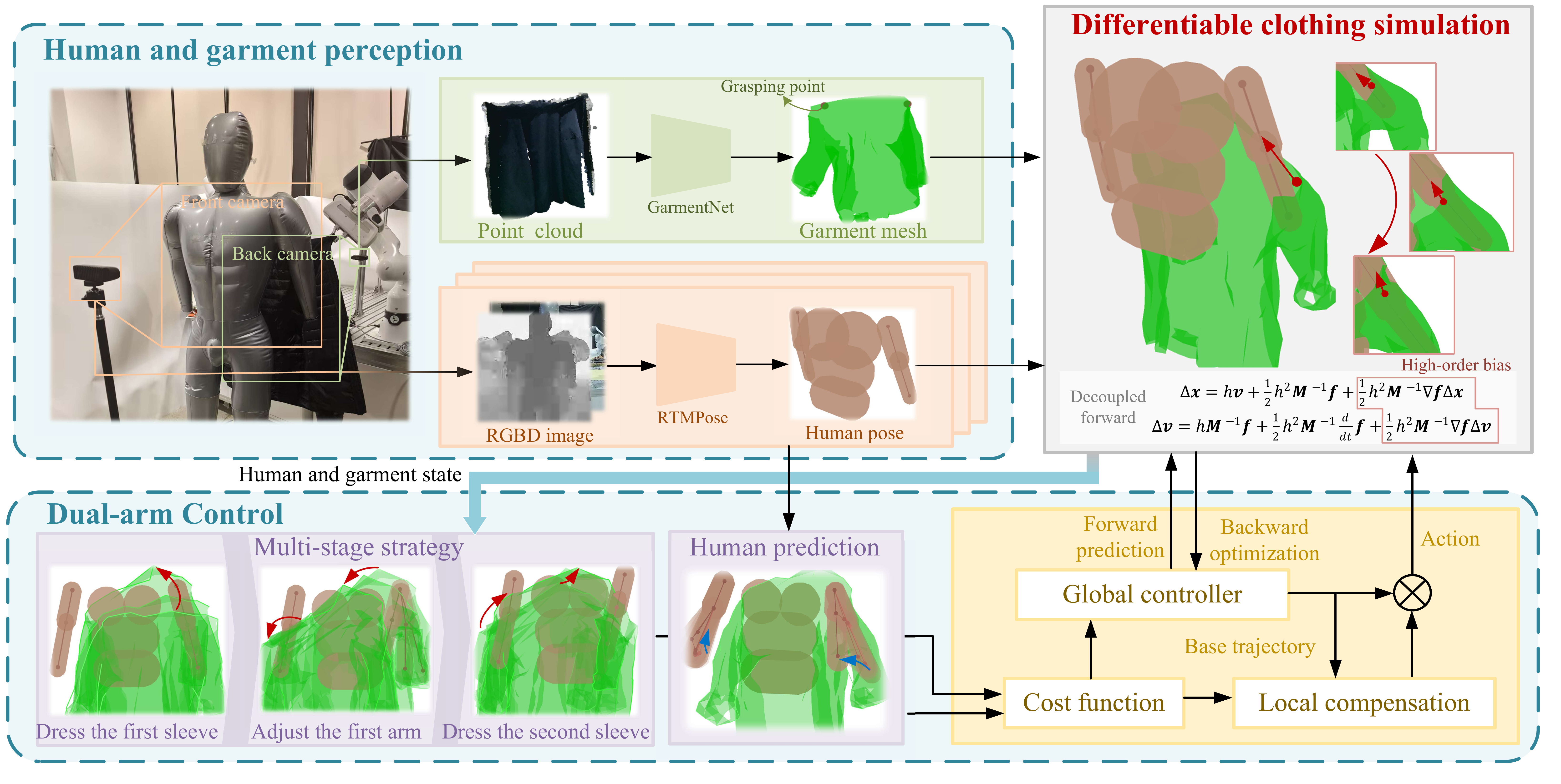}
        \caption{The pipeline of our proposed robotic-assisted dressing framework. The system comprises three components: human and garment perception, differentiable clothing simulation, and dual-arm control. The perception module provides the clothing reconstruction and human pose, facilitating the differentiable clothing simulation. Using the simulated human and garment states, the control module decomposes the dressing task into multiple sequential phases. By incorporating human prediction, the framework accomplishes the dressing procedure through an optimal control scheme with local compensation.}
        \label{pipeline}
\end{figure*}
To achieve robotic-assisted dressing, we propose an integrated framework as illustrated in Fig. \ref{pipeline}. This framework employs garment point clouds and human RGB-D images acquired from depth cameras to perform garment mesh reconstruction, tracking, and human pose estimation. 
Building upon the perception results, we develop a novel differentiable clothing simulation algorithm. This algorithm adopts a position-velocity decoupled explicit iterative scheme to enhance computational efficiency while incorporating high-order bias terms to ensure stability under large time steps. 
By leveraging the simulated human and garment states, our framework decomposes the dressing task into a sequential multi-stage process: dressing the first sleeve, adjusting the arm, and dressing the second sleeve. Coupled with human motion prediction, these stages are mathematically defined via cost functions, steering the dressing progression.
Ultimately, robotic dressing assistance is achieved through a simulation-based global optimal controller, augmented with real-time local compensation to facilitate efficient, rapid, and coordinated dual-arm control.

\section{Differentiable Clothing Simulation}
\label{sec:simulation}
\subsection{Forward Simulation} 
\label{subsec:Forward}
Denote $\boldsymbol{x}(t)=[\boldsymbol{p}_1^T,...,\boldsymbol{p}_n^T]^T$, $\boldsymbol{v}(t)=[\boldsymbol{\dot{p}}_1^T,...,\boldsymbol{\dot{p}}_n^T]^T$ as the positions and velocities of all nodes at time $t$. Denote $h$ as the time step and $\boldsymbol{M}\in\mathbb{R}^{3n\times 3n}$ as the positive diagonal mass matrix. $\boldsymbol{f}_{int}\in\mathbb{R}^{3n}, \boldsymbol{f}_{ext}\in\mathbb{R}^{3n}$ are the internal and external forces at all nodes. Based on the Taylor expansion, we consider the  time-stepping scheme discretizing $\boldsymbol{x}(t)$ and $\boldsymbol{v}(t)$ as follows:
\begin{align}
    \begin{split}
        &\boldsymbol{x}_{n+1} = \boldsymbol{x}_n + h\boldsymbol{v}_{n} + \frac{1}{2} h^2 \boldsymbol{M}^{-1}(\boldsymbol{f}_{ext} + \boldsymbol{f}_{int}) + ... \\ 
        &\boldsymbol{v}_{n+1} = \boldsymbol{v}_n + h\boldsymbol{M}^{-1}(\boldsymbol{f}_{ext} + \boldsymbol{f}_{int})  \\
         &\phantom{\boldsymbol{v}_{n+1} = \boldsymbol{v}_n }+ \frac{1}{2}h^2\boldsymbol{M}^{-1}\frac{d}{dt}(\boldsymbol{f}_{ext}+ \boldsymbol{f}_{int})+...
    \end{split} 
    \label{iteration}
\end{align}
Notably, in the following, we assume $\boldsymbol{f}_{int},\boldsymbol{f}_{ext}$ do not contain contact forces, which will be introduced in Sec. \ref{subsec:contact}. $\boldsymbol{f}_{ext}$ is calculated as the constant gravity and wind forces. $\boldsymbol{f}_{int}$ is calculated based on the cloth material model described in \cite{BouazizProjectiveDynamicsFusing2023}. 
\par
Referring to \cite{LiuQuasiNewtonMethods2017}, we can define the potential energy $W$ to calculate $\boldsymbol{f}_{int}$ by its spatial gradient: $\boldsymbol{f}_{int}= -\nabla W(\boldsymbol{x}_n)$. $W$ can be divided into a sum of quadratic terms:
\begin{align}
    W(\boldsymbol{x}_n) = \sum_i \min_{\boldsymbol{q}_i\in\mathcal{M}_i} \frac{w_i}{2}\|\boldsymbol{A}_i\boldsymbol{x}_n - \boldsymbol{q}_i\|^2_2 
    \label{Wx}
\end{align}
Each constraint corresponds to one quadratic term with stiffness $w_i$ and linear transformation operator $\boldsymbol{A}_i$. $\boldsymbol{q}_i$ is the auxiliary projection variable and $\mathcal{M}_i$ is the constraint manifold. More details about their definitions can be found in \cite{LiuQuasiNewtonMethods2017}. According to (\ref{Wx}), the derivative of $\boldsymbol{f}_{int}$ can be obtained as:
\begin{align}
    \frac{d}{dt} \boldsymbol{f}_{int}=-\nabla^2 W(\boldsymbol{x}_n) \frac{d\boldsymbol{x}_n}{dt} = \sum_i w_i(\boldsymbol{A}_i^T\nabla \boldsymbol{q}_i - \boldsymbol{A}_i^T\boldsymbol{A}_i)\boldsymbol{v}_n
    \label{df_dx}
\end{align}
Assuming elastic potential energy dominates the deformation behavior, the potential energy $W$ can be decomposed into stretching and bending components, whose constraint manifold $\mathcal{M}_i$ is $SO(3)$ with $\boldsymbol{q}_i\in\mathbb{R}^{3\times 3}$. Denote $\boldsymbol{\hat{A}}_i\boldsymbol{x}_n - \boldsymbol{\hat{q}}_i\in\mathbb{R}^{9}$ to expand $\boldsymbol{A}_i\boldsymbol{x}_n - \boldsymbol{q}_i$ as a vector. $\boldsymbol{\hat{q}}_i$ is calculated from a rotation matrix with three freedoms, whose differentiation can be defined as:
\begin{align}
    d\boldsymbol{\hat{q}}_i = \boldsymbol{B}_i(\boldsymbol{\hat{q}}_i) d\boldsymbol{\omega}_i
\end{align}
where $\boldsymbol{B}_i(\boldsymbol{\hat{q}}_i)\in \mathbb{R}^{9\times 3}$ is the Jacobian matrix with respect to $\boldsymbol{\omega}_i \in \mathbb{R}^3$. At the minimizer $\boldsymbol{\hat{q}}^*_i\in \mathbb{R}^{9}$ of $\boldsymbol{\hat{q}}_i \mapsto\|\boldsymbol{\hat{A}}_i\boldsymbol{x}_n - \boldsymbol{\hat{q}}_i\|^2$, the spatial gradient $\nabla \boldsymbol{\omega}_i$ satisfies:
\begin{align}
    \nabla \boldsymbol{\omega}_i = \boldsymbol{B}_i(\boldsymbol{\hat{q}}_i^*)^{\dagger}\boldsymbol{\hat{A}}_i
\end{align}
where $(\cdot)^{\dagger}$ is the Moore-Penrose pseudoinverse. Thus, the term $\boldsymbol{A}_i^T\nabla \boldsymbol{q}_i$ can be calculated as:
\begin{align}
    \boldsymbol{A}_i^T \nabla \boldsymbol{q}_i =\boldsymbol{\hat{A}}_i^T \boldsymbol{B}_i(\boldsymbol{\hat{q}}_i^*)\boldsymbol{B}_i(\boldsymbol{\hat{q}}_i^*)^{\dagger}\boldsymbol{\hat{A}}_i
    \label{dp_dx}
\end{align}
\par
The explicit iterative method using (\ref{iteration}) performs poorly and unstably under large steps and large stiffness. In this case, the internal force changes significantly at each iteration, which can be captured by Taylor expansion only in a brief interval. With the increase of iteration steps, the deviation of force makes the simulation oscillation and unstable. By contrast, the implicit Euler method \cite{BouazizProjectiveDynamicsFusing2023} calculates the force based on the next state $\boldsymbol{x}_{n+1}$, which incorporates the variation of force promoted by its own influence. However, the implicit method needs to solve the non-linear equation to obtain $\boldsymbol{x}_{n+1}$, which restricts the solving speed and affects the calculation of differentiation.
\par
To avoid the nonlinear equation and ensure fast solution, we introduce the force compensation into the second-order forward Taylor expansion as follows:
\begin{align}
    \begin{split}
        &\boldsymbol{x}_{n+1} = \boldsymbol{x}_{n} + h\boldsymbol{v}_{n} + \frac{1}{2} h^2 \boldsymbol{M}^{-1}(\boldsymbol{f}_{ext} + \boldsymbol{f}_{int}) + h^2\boldsymbol{F}_x \Delta \boldsymbol{x}  \\ 
        &\boldsymbol{v}_{n+1}= \boldsymbol{v}_{n} +  h\boldsymbol{M}^{-1} (\boldsymbol{f}_{ext} + \boldsymbol{f}_{int}) + \\
        &\phantom{\boldsymbol{v}_{n+1}= \boldsymbol{v}_{n} } \frac{1}{2} h^2 \boldsymbol{M}^{-1} \frac{d}{dt}(\boldsymbol{f}_{ext}+ \boldsymbol{f}_{int}) + h^2 \boldsymbol{F}_x \Delta \boldsymbol{v} \\
    \end{split} 
    \label{standard-iteration}
\end{align}
where $\Delta \boldsymbol{x} = \boldsymbol{x}_{n+1}-\boldsymbol{x}_n$, $\Delta \boldsymbol{v} = \boldsymbol{v}_{n+1}-\boldsymbol{v}_n$, and $\boldsymbol{F}_x =\boldsymbol{M}^{-1}\nabla \boldsymbol{f}_{int}$. Referred to the implicit Euler method, the compensation terms utilize the variation to calculate the change of internal force, which incorporates the impact of the state change and overcomes the instability of the explicit method. Notably, we introduce distinct compensation terms for position and velocity, thereby decoupling the consideration of nodal velocity and position changes. This approach accounts for scenarios involving large step sizes, where the nodal velocity may rapidly decay to zero in the initial phase of a single step, independent of the average velocity, as well as the displacement within that step.
Through repeated iterations, $\Delta \boldsymbol{x}$ and $\Delta \boldsymbol{v}$ can be written as:
\begin{align}
    \begin{split}
        &\Delta \boldsymbol{x} = \sum_{i=1}^\infty (h^2\boldsymbol{F}_x)^i [h\boldsymbol{v}_{n}  + \frac{1}{2} h^2 \boldsymbol{M}^{-1}(\boldsymbol{f}_{ext} + \boldsymbol{f}_{int})] \\
        &\Delta \boldsymbol{v} = \sum_{i=1}^\infty (h^2\boldsymbol{F}_x)^i (h\boldsymbol{M}^{-1} + \frac{1}{2} h^2 \boldsymbol{M}^{-1} \frac{d}{dt})  (\boldsymbol{f}_{ext} + \boldsymbol{f}_{int}) \\
    \end{split} 
    \label{delta-definiation}
\end{align}
The compensation terms satisfy $\Delta \boldsymbol{x} = o(h^3)$ and $\Delta \boldsymbol{v} = o(h^3)$, serving as high-order bias terms for both position and velocity in their respective Taylor expansions. For sufficiently small timestep $h$, our method preserves accuracy equivalent to second-order Taylor expansion. As the increase of the iterative steps, (\ref{delta-definiation}) diverges. Based on (\ref{standard-iteration}), $\Delta \boldsymbol{x}$ and $\Delta \boldsymbol{v}$ can be solved by linear equation, which can be written as:
\begin{align}
    \begin{split}
    &\Delta \boldsymbol{x} = (\boldsymbol{I} - h^2\boldsymbol{F}_x)^{-1} [h\boldsymbol{v}_{n} + \frac{1}{2} h^2 \boldsymbol{M}^{-1}(\boldsymbol{f}_{ext} + \boldsymbol{f}_{int})]\\
    &\Delta \boldsymbol{v} = (\boldsymbol{I} - h^2\boldsymbol{F}_x)^{-1} (h\boldsymbol{M}^{-1} +  \frac{1}{2} h^2 \boldsymbol{M}^{-1} \frac{d}{dt})  (\boldsymbol{f}_{ext} + \boldsymbol{f}_{int})
    \end{split} 
    \label{delta-definiation2}
\end{align}
According to (\ref{df_dx}) and (\ref{dp_dx}), we have:
\begin{align}
    \begin{split}
    (\boldsymbol{I} - &h^2\boldsymbol{F}_x)^{-1} = [\boldsymbol{M} +\\
     &h^2\sum_i w_i\boldsymbol{\hat{A}}_i^T(\boldsymbol{I} - \boldsymbol{B}_i(\boldsymbol{\hat{q}}_i^*)\boldsymbol{B}_i(\boldsymbol{\hat{q}}_i^*)^{\dagger})\boldsymbol{\hat{A}}_i]^{-1}\boldsymbol{M}^{-1}
    \end{split}
    \label{Ih2Fx}
\end{align}
As $(\boldsymbol{I} - \boldsymbol{B}_i(\boldsymbol{\hat{q}}_i^*)\boldsymbol{B}_i(\boldsymbol{\hat{q}}_i^*)^{\dagger})$ is a projection matrix, our force compensation can be seen as the semi-definite compensation on the mass matrix $\boldsymbol{M}$, which reduces the variation $\Delta \boldsymbol{x}$ and $\Delta \boldsymbol{v}$ in one step. Under large steps and large stiffness, $\Delta \boldsymbol{x}$ and $\Delta \boldsymbol{v}$ satisfy:
\begin{align}
    \begin{split}
        & \lim_{h,w_i->\infty}\Delta \boldsymbol{x} =-\frac{1}{2}\boldsymbol{F}_x^{-1}\boldsymbol{M}^{-1}(\boldsymbol{f}_{int} + \boldsymbol{f}_{ext})  \\
        & \lim_{h,w_i->\infty}\Delta \boldsymbol{v} =-\frac{1}{2}\boldsymbol{F}_x^{-1}\boldsymbol{M}^{-1}\frac{d}{dt}\boldsymbol{f}_{int} = -\frac{1}{2}\boldsymbol{v}_n\\
    \end{split} \\
\end{align}
Thus, with the increase of the step $h$, our method reduces the variation in one step to approach the equilibrium state with zero force to ensure stability. 
\par
To accelerate the calculation, we use iterative solver to replace the matrix inversion. The solution for $\Delta \boldsymbol{x}$ can be written as:
\begin{align}
    \begin{split}
        & \boldsymbol{P} \Delta \boldsymbol{x}_{k+1} = \Delta \boldsymbol{P} \Delta \boldsymbol{x}_{k} + h\boldsymbol{M}\boldsymbol{v}_{n} + \frac{1}{2} h^2 (\boldsymbol{f}_{ext} + \boldsymbol{f}_{int}) \\
        & \boldsymbol{P} = \boldsymbol{M} + h^2 \sum_i w_i\boldsymbol{A}_i^T\boldsymbol{A}_i \\
        & \Delta \boldsymbol{P} = h^2 \sum_i w_i\boldsymbol{A}_i^T\nabla \boldsymbol{q}_i \\
    \end{split}
    \label{iterative-solver}
\end{align}
where $\boldsymbol{P}$ is a fixed positive definite symmetric matrix, which can be decomposed by Cholesky factorization to accelerate the iteration. Such an iterative solver converges from any initialization when the spectral radius $\rho(\boldsymbol{P}^{-1}\Delta \boldsymbol{P})<1$. For any $\boldsymbol{z}\in \mathbb{R}^{3n}$, according to (\ref{dp_dx}), (\ref{iterative-solver}), we have:
\begin{align}
    \boldsymbol{z}^T\Delta \boldsymbol{P} \boldsymbol{z} \leq h^2 \sum_i w_i  \boldsymbol{z}^T\boldsymbol{A}_i^T\boldsymbol{A}_i\boldsymbol{z} < \boldsymbol{z}^T \boldsymbol{P} \boldsymbol{z}
\end{align}  
Thus, the Rayleigh Quotient for $\boldsymbol{P}^{-1}\Delta \boldsymbol{P}$ is less than $1$, as well as all the eigenvalues. The iterative solver converges for different constraints. $\Delta v$ can be derived similarly.
\par
In practical applications, the results of one-step iteration can be employed to achieve faster simulation. In this case, the accuracy is still guaranteed to be $o(h^3)$. Meanwhile, this approximation further reduces the variation of $\boldsymbol{x}$ and $\boldsymbol{v}$ as the Rayleigh Quotient for $\boldsymbol{P}^{-1}(\boldsymbol{P} -\Delta \boldsymbol{P})$ is less than $1$, making the new state closer to equilibrium in simulation. 
\par

\subsection{Contact Update}
\label{subsec:contact}
Signorini-Coulomb law describes the contacts force applied to each node respectively \cite{BernardNonsmoothLagrangianSystems2016}. At each step, denote the indices of all collision nodes as $\mathcal{I}\subseteq \{1,2,3,...n\}$.  $\mathcal{I}$ can be obtained as the mixture of the previous contact nodes and the new collision nodes. For each node $j\in\mathcal{I}$, its local contact force $\boldsymbol{r}_j\in \mathbb{R}^3$ and velocity $\boldsymbol{u}_j\in \mathbb{R}^3$ satisfy one of the following conditions:
\begin{align}
    \left\{\begin{aligned}
           & \text{Take off:} \phantom{12}\boldsymbol{r}_j =0, \boldsymbol{u}_j \cdot \boldsymbol{n}_j>0,\\
           & \text{Stick:}    \phantom{123w}||\boldsymbol{r}_{j|T}|| <\mu||\boldsymbol{r}_{j|N}||, \boldsymbol{u}_j =0,\\
           & \text{Slip:}     \phantom{1234Q}||\boldsymbol{r}_{j|T}|| = \mu ||\boldsymbol{r}_{j|N}||, \boldsymbol{u}_{j|N} =0, \boldsymbol{r}_{j|T} = -k_j\boldsymbol{u}_{j|T}
        \end{aligned}
    \right.
    \label{contact-constraint}
\end{align}
where $\mu$ is the frictional coefficient and $k_j$ is a positive coefficient. The notations $(\cdot)_{j|T}$ and $(\cdot)_{j|N}$ represent the tangential and normal components respect to the contact surface. $\boldsymbol{n}_j$ is the normal direction deviated from the surface for node $j$. With $m$ contact nodes, the vertical stacks of all contact forces and velocities are defined as $\boldsymbol{r}\in\mathbb{R}^{3m}$ and $\boldsymbol{u}\in\mathbb{R}^{3m}$, respectively.
\par
 For convenience, we utilize a constant contact force  $\boldsymbol{r}$ at each step, which satisfies the contact conditions in (\ref{contact-constraint}). The corresponding $\boldsymbol{u}$ is calculated as the mean velocity: $\boldsymbol{u} = \Delta \boldsymbol{x}/ h$. According to (\ref{delta-definiation2}), $\boldsymbol{u}$ can be obtained based on the force applied to the nodes. When utilizing one-step iteration for forward simulation, the relationship between $\boldsymbol{r}$ and $\boldsymbol{u}$ satisfies:
 \begin{align}
    \begin{split}
&    \boldsymbol{u} = \frac{1}{2}h\boldsymbol{P}_{ic} \boldsymbol{r} + \boldsymbol{b}\\
& \boldsymbol{b} = \boldsymbol{I}_c \boldsymbol{P}^{-1}[\boldsymbol{M}\boldsymbol{v}_n + \frac{1}{2}h(\boldsymbol{f}_{ext} + \boldsymbol{f}_{int})]
    \end{split}
    \label{P-inverse}
 \end{align}
 where $\boldsymbol{P}_{ic}\in\mathbb{R}^{3m\times 3m}$ is the sub matrix by extracting the related rows and columns from $\boldsymbol{P}^{-1}$. $\boldsymbol{I}_c\in\mathbb{R}^{3m\times3n}$ is the matrix to extract the velocity of contact nodes from all nodes. $\boldsymbol{b}$ represents the motion trend without contact.
When performing a complete forward simulation, we can derive from (\ref{delta-definiation2}) a linear constraint relationship between $\boldsymbol{u}$ and $\boldsymbol{r}$, analogous in form to the expression given in (\ref{P-inverse}).
\par
To distinguish the contact types, we suppose all contact nodes stick and calculate the corresponding contact force as $-2\boldsymbol{P}_{ic}^{-1}\boldsymbol{b}/h$. Three types is distinguished by the following conditions:
\begin{align}
        \left\{\begin{aligned}
           & \text{Take off:} \phantom{12}\boldsymbol{r}_j\cdot \boldsymbol{n}_j< 0,\\
           & \text{Stick:}    \phantom{123w}||\boldsymbol{r}_{j|T}|| \leq\mu||\boldsymbol{r}_{j|N}||,\boldsymbol{r}_j\cdot \boldsymbol{n}_j> 0, \\
           & \text{Slip:}     \phantom{1234Q}||\boldsymbol{r}_{j|T}|| >\mu ||\boldsymbol{r}_{j|N}||, \boldsymbol{r}_j\cdot \boldsymbol{n}_j> 0
        \end{aligned}
    \right.
    \label{contact-type}
\end{align}
\par
\begin{figure}[!t]\centering
    \includegraphics[width=0.48\textwidth,trim=0 0 0 0,clip]{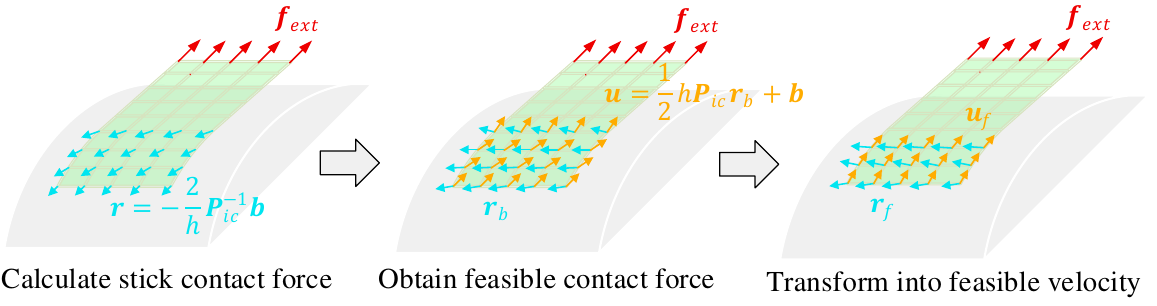}
    \caption{An example of the process to solve the contact force: A rectangular fabric patch is placed on a cylindrical surface and subjected to a horizontal external force $\boldsymbol{f}_{ext}$. The fabric exhibits sliding motion with edge detachment. The figure demonstrates the calculation of the sticking force, the feasible force $\boldsymbol{r}_{b}$ and the resulting force $\boldsymbol{r}_f$ with feasible velocity $\boldsymbol{u}_f$.}
    \label{contact_process}
\end{figure}

Based on the contact types, we define the matrix $\boldsymbol{C}_r\in\mathbb{R}^{3m\times 3m}$, $\boldsymbol{C}_u\in\mathbb{R}^{3m\times 3m}$ to apply the constraints in (\ref{contact-constraint}) for $\boldsymbol{r}$ and $\boldsymbol{u}$, respectively. The direction constraint $\boldsymbol{r}_{j|T} = -k_j\boldsymbol{u}_{j|T}$ for slipping nodes is ignored as our contact force and velocity are the mean values for process and do not correspond exactly. Then, the feasible contact force $\boldsymbol{r}_{b}\in\mathbb{R}^{3m}$ to prevent the motion $\boldsymbol{b}$ is obtained as:
\begin{align}
\boldsymbol{r}_{b} = -\frac{2}{h}\boldsymbol{C}_r\boldsymbol{P}_{ic}^{-1}\boldsymbol{b}
\end{align}
However, $\boldsymbol{u}$ corresponding to this contact force may violate (\ref{contact-constraint}). For the process values, the nodal velocity satisfies the constraints always to ensure they remain in the surface while the force does not need. Thus, we introduce $\boldsymbol{r}_b$ into (\ref{P-inverse}) to obtain the feasible velocity $\boldsymbol{u}_f\in\mathbb{R}^{3m}$ and the corresponding force $\boldsymbol{r}_f\in\mathbb{R}^{3m}$:
\begin{align}
    \begin{split}
        &\boldsymbol{u}_f = \boldsymbol{C}_u(\boldsymbol{I}-\boldsymbol{P}_{ic}\boldsymbol{C}_r\boldsymbol{P}_{ic}^{-1})\boldsymbol{b}\\
        &\boldsymbol{r}_f = \frac{2}{h}\boldsymbol{P}_{ic}^{-1} (\boldsymbol{u}_f - \boldsymbol{b})
    \end{split}
    \label{obtain-rf}
\end{align}
A representative example of the solution process is illustrated in Fig. \ref{contact_process}, in which a rectangular fabric patch is slid and partially detached on a cylindrical surface under applied tension.
\par
Thus, the fast explicit iteration with contact force can be written as:
\begin{align}
    \begin{split}
    &\boldsymbol{x}_{n+1} = \boldsymbol{x}_n + \boldsymbol{P}^{-1} [h\boldsymbol{M}\boldsymbol{v}_{n} + \frac{1}{2} h^2 (\boldsymbol{f}_{ext} + \boldsymbol{f}_{int}+\boldsymbol{I}_c^T\boldsymbol{r}_f)]\\
    &\boldsymbol{v}_{n+1} = \boldsymbol{v}_n + \boldsymbol{P}^{-1}(h+ \frac{1}{2} h^2 \frac{d}{dt})  (\boldsymbol{f}_{ext} + \boldsymbol{f}_{int}+\boldsymbol{I}_c^T\boldsymbol{r}_f)
    \end{split} 
    \label{force-contact}
\end{align}
\par
\textit{Remark:} Considering that the object movement and uneven contact surfaces may cause the contact nodes following (\ref{contact-constraint}) to detach, we make corrections to all contact points after collision detection. The contact nodes are first updated to ensure their positions unchanged in the object coordinate system. Then, we project them to the nearest positions on contact surfaces to maintain contact.
\par
In summary, our method utilizes the advantage of explicit iteration to construct the relationship between contact force and new state, which ensures the friction constraints and quick calculation. The update process is shown as Algorithm \ref{forward-update}.
\begin{algorithm}
    \SetAlgoLined
    \KwData{Mass matrix $\boldsymbol{M}$, time step $h$, total steps $N_s$, potential energy function $W$, initial nodal position $\boldsymbol{x}_0$ and velocity $\boldsymbol{v}_0$} 
    \KwIn{external force  $\boldsymbol{f}_{ext}$}
    $\mathcal{I} \gets \emptyset, n\gets 0$\;
    $\boldsymbol{P} \gets \boldsymbol{M} + h^2\sum_i w_i \boldsymbol{A}_i^T\boldsymbol{A}_i $\;
    pre-calculate $\boldsymbol{P}^{-1}$\;
    \While{$n<N_s$}{
        $n\gets n+1$\;
        get external force $\boldsymbol{f}_{ext}$\; 
        $\mathcal{I} \gets \mathcal{I} \cup CheckCollision(\boldsymbol{x}_n)$\;
        $\boldsymbol{x}_n \gets CorrectContact(\mathcal{I}, \boldsymbol{x}_n)$\;
        $\boldsymbol{f}_{int} \gets -\nabla W(\boldsymbol{x}_n)$\;
        extract $\boldsymbol{P}_{ic}$ from $\boldsymbol{P}^{-1}$\; 
        calculate $\boldsymbol{b}$ by (\ref{P-inverse})\;
        $\boldsymbol{r} \gets -2\boldsymbol{P}_{ic}^{-1}\boldsymbol{b}/h$\;
        distinguish contact types by (\ref{contact-type})\;
        calculate $\boldsymbol{r}_f$ by (\ref{obtain-rf})\; 
        calculate $\boldsymbol{x}_{n+1}, \boldsymbol{v}_{n+1}$ by (\ref{force-contact})\;
        update $\mathcal{I}$
    }
\caption{Explicit Iteration with Contact Force}
\label{forward-update}
\end{algorithm}

\subsection{Backward Differentiation}
\label{subsec:backward}
The key idea to achieve differentiation is to derive the Jacobian of the output $(\boldsymbol{x}_{n+1}, \boldsymbol{v}_{n+1})$ with respect to the input $(\boldsymbol{x}_n, \boldsymbol{v}_n)$. Regarding the forward iteration (\ref{iterative-solver}) without contact, we can differentiate the equivalent formula (\ref{standard-iteration}) to obtain:
\begin{align}
    \begin{split}
        &d\boldsymbol{x}_{n+1} = d\boldsymbol{x}_n + hd\boldsymbol{v}_n + \frac{1}{2}h^2\boldsymbol{F}_xd\boldsymbol{x}_n + 
        h^2\boldsymbol{F}_x(d\boldsymbol{x}_{n+1} \\
        & \phantom{123456789} - d\boldsymbol{x}_n) + h^2\nabla \boldsymbol{F}_x \Delta \boldsymbol{x} d\boldsymbol{x}_n \\
        &d\boldsymbol{v}_{n+1} = d\boldsymbol{v}_n + h\boldsymbol{F}_xd\boldsymbol{x}_n +\frac{1}{2}h^2\boldsymbol{F}_x d\boldsymbol{v}_n +\frac{1}{2}h^2\nabla \boldsymbol{F}_x  \boldsymbol{v}_nd\boldsymbol{x}_n\\
        &  \phantom{123456789}  + h^2\boldsymbol{F}_x(d\boldsymbol{v}_{n+1} - d\boldsymbol{v}_n) + h^2\nabla \boldsymbol{F}_x \Delta \boldsymbol{v}d\boldsymbol{x}_n
    \end{split}
\end{align}
which can be re-written as:
\begin{align}
    \begin{split}
        &\begin{bmatrix}
        \boldsymbol{I} - h^2\boldsymbol{F}_x &  0\\
        0 & \boldsymbol{I} - h^2\boldsymbol{F}_x
    \end{bmatrix}
    \begin{bmatrix}
        d\boldsymbol{x}_{n+1}\\
        d\boldsymbol{v}_{n+1}
    \end{bmatrix} = \boldsymbol{H} \begin{bmatrix}
        d\boldsymbol{x}_{n}\\
        d\boldsymbol{v}_{n}
    \end{bmatrix}
\end{split}
\label{diff-delta}
\end{align}
where 
\begin{align}
    \boldsymbol{H} = \begin{bmatrix}
        \boldsymbol{I} -\frac{1}{2}h^2\boldsymbol{F}_x +h^2\nabla \boldsymbol{F}_x \Delta x & h\boldsymbol{I} \\
        h\boldsymbol{F}_x + h^2\nabla \boldsymbol{F}_x (\frac{1}{2}\boldsymbol{v}_n + \Delta \boldsymbol{v})  &\boldsymbol{I} -\frac{1}{2}h^2\boldsymbol{F}_x 
    \end{bmatrix}
\end{align}
\par
In back propagation, the Jacobian matrices are coupled with the gradients of a loss function $L$. We can pass the gradient vector with respect to $L$ to replace the Jacobian matrix. By utilizing the chain role to (\ref{diff-delta}), we have:
\begin{align}
    \begin{bmatrix}
    \frac{\partial L}{\partial \boldsymbol{x}_{n}}\\
    \frac{\partial L}{\partial \boldsymbol{v}_{n}}    
    \end{bmatrix} = \boldsymbol{H}^T 
        \begin{bmatrix}
    (\boldsymbol{I} - h^2\boldsymbol{F}^T_x)^{-1}\frac{\partial L}{\partial \boldsymbol{x}_{n+1}}\\
    (\boldsymbol{I} - h^2\boldsymbol{F}^T_x)^{-1}\frac{\partial L}{\partial \boldsymbol{v}_{n+1}}  
    \end{bmatrix}
\end{align}
Denote $\boldsymbol{d}= (\boldsymbol{I} - h^2\boldsymbol{F}^T_x)^{-1}{\partial L}/{\partial \boldsymbol{x}_{n+1}}$. $\boldsymbol{d}$ can be calculated based on the iterative method similar to (\ref{iterative-solver}):
\begin{align}
    \boldsymbol{P}\boldsymbol{M}^{-1}\boldsymbol{d}_{k+1} = \Delta \boldsymbol{P}^T \boldsymbol{M}^{-1}\boldsymbol{d}_k + \frac{\partial L}{\partial \boldsymbol{x}_{n+1}}
\end{align}
$(\boldsymbol{I} - h^2\boldsymbol{F}^T_x)^{-1}{\partial L}/{\partial \boldsymbol{v}_{n+1}}$ is calculated in the same way. The number of iterations is the same as the forward update to ensure consistency in the solution. The iteration is converged as verified in Sec \ref{subsec:Forward}.
\par
Under contact, the forward iteration is solved as (\ref{force-contact}). The contact force and the new states are calculated through matrix multiplication directly. We can take the derivative to obtain the 
directly calculated through matrix multiplication. By Differentiating $\boldsymbol{x}_n$ and $\boldsymbol{v}_n$ on both sides respectively, we have:
\begin{align}
    \begin{split}
        &\frac{\partial \boldsymbol{x}_{n+1}}{\partial \boldsymbol{x}_n} = \boldsymbol{I} + \frac{1}{2} h^2\boldsymbol{P}^{-1}(\frac{\partial\boldsymbol{f}_{int}}{\partial \boldsymbol{x}_n} +\boldsymbol{I}_c^T\frac{\partial \boldsymbol{r}_f}{\partial \boldsymbol{x}_n})\\
        &\frac{\partial \boldsymbol{x}_{n+1}}{\partial \boldsymbol{v}_n} =  h\boldsymbol{P}^{-1}\boldsymbol{M} +\frac{1}{2} h^2\boldsymbol{P}^{-1}\boldsymbol{I}_c^T\frac{\partial \boldsymbol{r}_f}{\partial \boldsymbol{v}_n}\\
        &\frac{\partial \boldsymbol{v}_{n+1}}{\partial \boldsymbol{x}_n} = h\boldsymbol{P}^{-1}(\frac{\partial\boldsymbol{f}_{int}}{\partial \boldsymbol{x}_n} +\boldsymbol{I}_c^T\frac{\partial \boldsymbol{r}_f}{\partial \boldsymbol{x}_n}+ \frac{1}{2} h \frac{\partial^2\boldsymbol{f}_{int}}{\partial \boldsymbol{x}^2_n} \boldsymbol{v}_n) \\
        &\frac{\partial \boldsymbol{v}_{n+1}}{\partial \boldsymbol{v}_n} =\boldsymbol{I} + h\boldsymbol{P}^{-1}(\boldsymbol{I}_c^T\frac{\partial \boldsymbol{r}_f}{\partial \boldsymbol{v}_n}+ \frac{1}{2} h \frac{\partial\boldsymbol{f}_{int}}{\partial \boldsymbol{x}_n}) \\
    \end{split}
    \label{backward-update}
\end{align}
According to  (\ref{P-inverse}) and (\ref{obtain-rf}), the derivative of $\boldsymbol{r}_f$ is:
\begin{align}
    \begin{split}
        &\frac{\partial \boldsymbol{r}_f}{\partial \boldsymbol{x}_n} = \boldsymbol{C}_p \frac{\partial \boldsymbol{f}_{int}}{\partial \boldsymbol{x}_n} \\
        &\frac{\partial \boldsymbol{r}_f}{\partial \boldsymbol{v}_n} = \frac{2}{h}\boldsymbol{C}_p\boldsymbol{M} \\
    \end{split}
\end{align}
where $\boldsymbol{C}_p = \boldsymbol{P}_{ic}^{-1}(\boldsymbol{C}_u - \boldsymbol{C}_u\boldsymbol{P}_{ic}\boldsymbol{C}_r\boldsymbol{P}_{ic}^{-1} - \boldsymbol{I})\boldsymbol{I}_c\boldsymbol{P}^{-1}$.  For convenience, the derivative is written as $\boldsymbol{J}_{f_{sim}}$ in the following.

\section{Multi-Stage Dual-Arm Control}
\label{sec:control}
\subsection{Task Decomposition} 
Based on the differentiable clothing simulation, the model predictive control is employed to robot control. Building upon human motion prediction, the bimanual optimization addresses human-robot interaction dynamics and inter-manipulator coordination for garment dressing tasks. Denote the current sleeve states as $\boldsymbol{x}_{ci,0}, i\in\{l,r\}$. Following the robot motion $\boldsymbol{v}_{ri,j} = \boldsymbol{p}_{ri,j+1} - \boldsymbol{p}_{ri,j}, i\in\{l,r\}$ and predicted human trajectory $\boldsymbol{p}_{x,j}, x\in\{sl,sr,el,er,hl,hr\}$ with time interval $h$, the predicted states $\boldsymbol{x}_{ci,j}, i\in\{l,r\}$ can be calculated through the simulation. Due to the motion constraints between two sleeves and direction limitation for insertion, we decompose the dressing task into multiple stages. The objective function $\mathcal{J}_{task}$ in the predictive horizon $N$ for different stages is defined respectively as follows.
\par
For one sleeve case, to ensure that the sleeve can fit smoothly onto the arm, we divide the dressing process into two stages: pulling the armhole to the hand and putting the sleeve up. In the beginning, we try to pull the armhole $\boldsymbol{p}_{ci}$ behind the hand, with the sleeve direction $\boldsymbol{v}_{ci}$ pointing towards the hand $\boldsymbol{p}_{hi}$. By dragging the sleeves to the back, the clothing can avoid contact with the user before dressing, providing convenience for later insertion. With the front direction of the human denoted as $\boldsymbol{v}_f$, the cost function $\mathcal{J}_{pi}$ for one sleeve is given as:
\begin{align}
    \begin{split}
    &\mathcal{J}_{pi}(\boldsymbol{x}_{ci}, \boldsymbol{v}_{ri}) = \sum_{j=0}^{N-1}(||\boldsymbol{p}_{hi,j} - l_h\boldsymbol{v}_f - \boldsymbol{p}_{ci,j}||_W^2 + \\
    &\phantom{1235111111155511115551}||\boldsymbol{v}_{ci,j} - \boldsymbol{v}_f||_Q^2 + \|\boldsymbol{v}_{ri,j}\|_R^2)
    \end{split}
    \label{Jpi}
\end{align}
where $\|\cdot\|_W^2$ stands for the square norm weighted by $\boldsymbol{W}$. $\boldsymbol{W}$, $\boldsymbol{Q}$ and $\boldsymbol{R}$ are positive diagonal weighting matrices. $l_h$ is a varied scalar to control the distance between the sleeve and the hand, which is defined as:
\begin{align}
    l_h = \left\{ 
        \begin{aligned}
        &\bar{l}_h, \phantom{234}\text{ If } \sqrt{\|(\boldsymbol{I}- \boldsymbol{v}_f\boldsymbol{v}_f^T)(\boldsymbol{p}_{hi,j} - \boldsymbol{p}_{ci,j})\|}>\bar{l}_h \\
        &\sqrt{\|(\boldsymbol{I}- \boldsymbol{v}_f\boldsymbol{v}_f^T)(\boldsymbol{p}_{hi,j} - \boldsymbol{p}_{ci,j})\|}, \phantom{234} \text{ o.w. }
        \end{aligned}
    \right.
\end{align}
$l_h$ gradually decreases as the sleeve approaches the hand, causing the armhole to eventually reach the hand position.
\par
When the cost $\mathcal{J}_{pi}$ is the less than the threshold $\phi_p$, the sleeve is prepared and the operation process enters the second stage. Denote the current armhole in the dressing system as $(s_i, l_i, \theta_i)$. The cost function $\mathcal{J}_{di}$ in this stage is defined as:
\begin{align}
    \begin{split}
    &\mathcal{J}_{di}(\boldsymbol{x}_{ci}, \boldsymbol{v}_{ri}) = \sum_{j=0}^{N-1}(||\boldsymbol{p}_{ci,j} - \mathcal{G}_{i,j}(s_i + j k)||_W^2 + \\
    &\phantom{1111111111111}||\boldsymbol{v}_{ci,j} - \mathcal{D}_{i,j}(s_i + j k)||_Q^2 + || \boldsymbol{v}_{ri,j}||_R^2)
    \end{split}
    \label{Jdi}
\end{align}
where $\mathcal{G}_{i,j}(s_i)$ and $\mathcal{D}_{i,j}(s_i)$ are the world coordinate and  arm direction calculated at the arm coordinate $(s_i, 0, 0)$, respectively. $s_i$ is limited in $[0, 2]$ before calculation. $k$ is the pre-defined distance for one step. As $s_i$ increases to $2$, assisted dressing with one single sleeve is completed.
\par
When considering two sleeves as a whole, the sleeves are restricted by the middle fabric connection and move with the maximum distance limit. This requires additional dressing policy to avoid clothes getting stuck between two arms as a local optima. Denote the sets of sleeves in the first stage and second stage as $\Lambda_p$ and $\Lambda_d$, respectively. Denote the index of the sleeve nearer to the arm as $n$ and the other as $f$. When the distance between two armholes is less than the threshold $\phi_c$, there is no restriction between two sleeves, and we obtain $\boldsymbol{v}_{rl}, \boldsymbol{v}_{rr}$ by applying the above loss functions for two sleeves, respectively. Otherwise, we deal with two sleeves one by one based their stages. $\phi_c$ is derived as the geodesic distance between two armholes from the clothing model.
\par
When $n,f\in\Lambda_p$, we prioritize putting sleeve $n$ on the arm, which is solved utilizing $\mathcal{J}_{pn}$ directly. Then, $n \in \Lambda_d$ while $f \in \Lambda_p$. In this configuration, the garment is constrained by the dressed arm. Upward pulling of sleeve $n$ potentially interferes with the donning process of the remaining parts. We proceed with donning sleeve $f$ while maintaining sleeve $n$ on arm. To minimize the restriction of sleeve $n$, we optimize the state $\boldsymbol{p}_{cn}$ to approach the other target hand $\boldsymbol{p}_{hf}$ along the arm. Let $(\hat{s}_f, \hat{l}_f, \hat{\theta}_f)$ represent the dressing coordinate of $\boldsymbol{p}_{hf}$ on the dressed arm. The cost function $\mathcal{J}_{wn}$ for sleeve $n$ is then formulated as:
\begin{align}
    \begin{split}
    &\mathcal{J}_{wn}(\boldsymbol{x}_{cn}, \boldsymbol{v}_{rn}) = \sum_{j=0}^{N-1}(||\boldsymbol{p}_{cn,j} - \mathcal{G}_{n,j}(\hat{s}_{f,j})||_W^2 + \\
    &\phantom{1111111111111111}||\boldsymbol{v}_{cn,j} - \mathcal{D}_{n,j}(\hat{s}_{f,j})||_Q^2 + || \boldsymbol{v}_{rn,j}||_R^2)
    \end{split}
    \label{Jwn}
\end{align}
To avoid detaching, we limit $\hat{s}_f$ to be larger than a positive threshold. $\boldsymbol{v}_{rl}, \boldsymbol{v}_{rr}$ are derived by combining the cost function $\mathcal{J}_{{wn}}$ and $\mathcal{J}_{pf}$. When $n,f\in\Lambda_d$, we obtain $\boldsymbol{v}_{rl}, \boldsymbol{v}_{rr}$ just like the unconstrained case, as two sleeves require to be pulled up both.
\par
Thus, the loss function $\mathcal{J}_{task}$ for different cases can be integrated as:
\begin{align}
    \mathcal{J}_{task} = \left\{
        \begin{aligned}
            &  \mathcal{J}_{\phi(l)l} + \mathcal{J}_{\phi(r)r},  \text{ If } \|\boldsymbol{p}_{cl} - \boldsymbol{p}_{cr}\| < \phi_c \lor \|\Lambda_p\| = 0 \\
            &  \mathcal{J}_{pn}, \phantom{2_12_s1_s2_s2_s} \text{ If } \|\boldsymbol{p}_{cl} - \boldsymbol{p}_{cr}\| \geq \phi_c \land \|\Lambda_p\| = 2 \\
            &  \mathcal{J}_{wn} + \mathcal{J}_{pf},\phantom{2_{234}} \text{ If } \|\boldsymbol{p}_{cl} - \boldsymbol{p}_{cr}\| \geq \phi_c \land \|\Lambda_p\| = 1  \\
        \end{aligned} 
        \right.
\end{align} 
where $\phi(i) \in \{p, d\}$ represents the stage of sleeve $i$.

\subsection{Model Predictive Control with Local Compensation}
\begin{figure}[!t]\centering
    \includegraphics[width=0.45\textwidth,trim=0 0 0 0,clip]{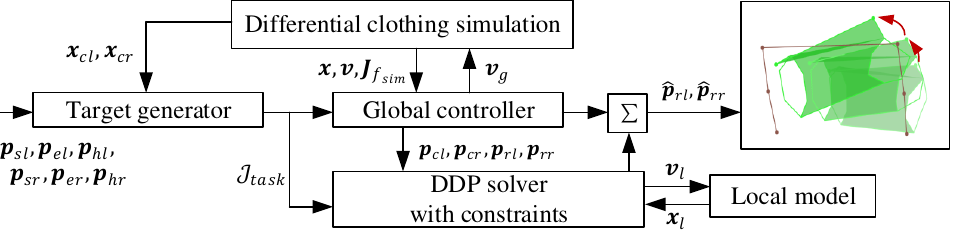}
    \caption{The control strategy designed for the dressing task.}
    \label{controller_frame}
\end{figure}

According to the loss function $\mathcal{J}_{task}$ obtained from task decomposition, we employ the differentiable simulation to calculate the global control quantity $\boldsymbol{v}_{g} =(\boldsymbol{v}_{rl}, \boldsymbol{v}_{rr})$. Meanwhile, due to the high computational costs involved in collision detection and fabric simulation, a simplified local model is utilized to compensate for $\boldsymbol{v}_g$ in real-time, which further improves the controller update speed. The controller framework is shown in the Fig. \ref{controller_frame}.
\par
When solving based on the global model, we introduce additional safety loss $\mathcal{J}_{safe}$. Commonly, the force between the clothing and human is limited to ensure wearing clothes safely. However, the contact area constantly changes during the dressing process, making it difficult to obtain gradients for optimization. Alternately, we control the actuator force, which can be seen as the combined force between the clothing and user, to ensure safe. Thus, $\mathcal{J}_{safe}$ is defined as: 
\begin{align}
    \mathcal{J}_{safe} = w_s\sum_{j=0}^{N-1}\sum_{i\in\Lambda_a}\frac{w_i}{2} \min_{\boldsymbol{q}_i\in\mathcal{M}_i}\|\boldsymbol{A}_i\boldsymbol{x}_j-\boldsymbol{q}_i\|_2^2
    \label{Jsafe}
\end{align}
where $\Lambda_a$ is the indices set of the actuator attachment constraints. $w_s$ is a positive weight coefficient. In this way, the global MPC law is obtained based on the following optimization problem:
\begin{align}
    \begin{split}
         \boldsymbol{v}_g &= \underset{\boldsymbol{v}_g}{\text{argmin}} \phantom{12} \mathcal{J}_{task} + \mathcal{J}_{safe}\\ 
         \text{subject to} \phantom{13} &\boldsymbol{x}_{j+1}, \boldsymbol{v}_{j+1} = f_{sim}(\boldsymbol{x}_j, \boldsymbol{v}_{j}, \boldsymbol{p}_{rl,j+1}, \boldsymbol{p}_{rr,j+1}) \\
        & \boldsymbol{p}_{rl,j+1} = \boldsymbol{p}_{rl,j} + \boldsymbol{v}_{rl,j}h\\
        & \boldsymbol{p}_{rr,j+1} = \boldsymbol{p}_{rr,j} + \boldsymbol{v}_{rr,j}h\\
        & \boldsymbol{x}_{cl,j}, \boldsymbol{x}_{cr,j} = f_{sleeve}(\boldsymbol{x}_j) \\
    \end{split}
    \label{global_mpc}
\end{align}
where $f_{sim}$ is the process of differentiable clothing simulation while $f_{sleeve}$ is the function to extract the sleeve state from the clothing nodes. Based on $\boldsymbol{J}_{f_{sim}}$, the derivative of $f_{sim}$ calculated in Sec. \ref{subsec:backward}, (\ref{global_mpc}) is solved by gradient descent method \cite{KirkOptimalControlTheory2004}.
\par
Due to the high dimension of the global model, the matrix operation in fabric simulation requires millisecond level time consumption. The point by point collision detection and the calculation of the inverse matrix $\boldsymbol{P}_{ic}$ in contact update further intensify the time required. Nonlinear optimization (\ref{global_mpc}) takes several seconds to complete, making our method unable to respond to human behaviors quickly and amplifying the impact of simulation errors. Thus, a linear local model is employed to compensated $\boldsymbol{v}_g$ and provide robot's real-time response to interaction. Denote the compensation for the sleeve states and control inputs as $\boldsymbol{x}_{l} =(\Delta \boldsymbol{p}_{cl}, \Delta \boldsymbol{p}_{cr}, \Delta \boldsymbol{p}_{rl}, \Delta \boldsymbol{p}_{rr})$ and $\boldsymbol{v}_{l} = (\Delta \boldsymbol{v}_{rl}, \Delta \boldsymbol{v}_{rr})$, respectively.
\par
In the local model, we directly consider the impact of robots on the sleeves states. When $\|\boldsymbol{p}_{cl} - \boldsymbol{p}_{cr}\|<\phi_c$, the change $\Delta\boldsymbol{p}_{ri}$ of end-effector only affects the corresponding sleeve $\Delta \boldsymbol{p}_{ci}$. In this case, due to the influence of gravity and contact force, the fabric between the $\boldsymbol{p}_{ci}$ and $\boldsymbol{p}_{ri}$ is straightened, making $\boldsymbol{p}_{ci}$ follow a similar motion to $\boldsymbol{p}_{ri}$. Thus, the compensation $\boldsymbol{x}_{l}$ is updated as:
\begin{align}
    \boldsymbol{x}_{l,j+1} = \boldsymbol{x}_{l,j} + (\boldsymbol{v}_{l,j},\boldsymbol{v}_{l,j})h
    \label{loc-func1}
\end{align} 
When $\|\boldsymbol{p}_{cl} - \boldsymbol{p}_{cr}\| \ge \phi_c$, $\Delta\boldsymbol{p}_{ri}$ affects both sleeves to keep their distance less than the threshold $\phi_f$, a geodesic distance from one armhole to the far pick point. The local model is defined as:
\begin{align}
    \boldsymbol{x}_{l,j+1} = \boldsymbol{x}_{l,j} + \left(\begin{bmatrix}
        1 - k_l & k_l \\
        k_r & 1 - k_r
    \end{bmatrix} \boldsymbol{v}_{l,j}, \boldsymbol{v}_{l,j}\right) h
    \label{loc-func2}
\end{align}
where
\begin{align}
    \begin{split}
        & k_l = \frac12 - \frac{\phi_f - \|\boldsymbol{p}_{cl,0} - \boldsymbol{p}_{rr,0}\|}{2(\phi_f - \phi_c)}\\
        & k_r = \frac12 - \frac{\phi_f - \|\boldsymbol{p}_{cr,0} - \boldsymbol{p}_{rl,0}\|}{2(\phi_f - \phi_c)}\\
    \end{split}
\end{align}
$k_l, k_r$ are influence weights that increase with distance. When $\|\boldsymbol{p}_{rl}-\boldsymbol{p}_{cr}\|$ or $\|\boldsymbol{p}_{rr}-\boldsymbol{p}_{cl}\|$ equal $\phi_f$, the corresponding weight reaches its maximum value of 1/2, indicating that the sleeve is equally affected by two grippers. 
\par
Denote the states with compensation as $\boldsymbol{\hat{p}}_{ci,j} = \boldsymbol{p}_{ci,j} + \Delta \boldsymbol{p}_{ci,j}$ and $\boldsymbol{\hat{p}}_{ri,j} = \boldsymbol{p}_{ri,j} + \Delta \boldsymbol{p}_{ri,j}$. A simple linear model cannot avoid the occurrence of infeasible states, e.g. $\|\boldsymbol{p}_{cl} - \boldsymbol{p}_{cr} \| > \phi_c$. Thus, we introduce additional constraints on the sleeve states $\boldsymbol{\hat{p}}_{ci}$ to represent the motion limitations from clothing stretching and external contact. To avoid excessive deformation, we limit the maximum distance between $\boldsymbol{\hat{p}}_{ci}$ and other points, which is written as:
\begin{align}
    \begin{aligned}
        &\|\boldsymbol{\hat{p}}_{cl,j} - \boldsymbol{\hat{p}}_{rr,j}\| \le \phi_f\\
        &\|\boldsymbol{\hat{p}}_{cr,j} - \boldsymbol{\hat{p}}_{rl,j}\| \le \phi_f\\
        &\|\boldsymbol{\hat{p}}_{cl,j} - \boldsymbol{\hat{p}}_{cr,j}\| \le \phi_c\\
    \end{aligned}
\end{align}
When the sleeve covers the arm as $i\in\Lambda_d$, the relative tangential position and speed between $\boldsymbol{\hat{p}}_{ci}$ and its target point $\mathcal{G}(s_i)$ are restricted to prevent the movement of the contact. This can be expressed as:
\begin{align}
    \begin{aligned}
        &\|\boldsymbol{D}_{i,j}(\boldsymbol{\hat{p}}_{ci,j} - \mathcal{G}_{i,j}(s_i + j k))\| \le \phi_x \quad \lor i \in \Lambda_p \\
        &\|\boldsymbol{D}_{i,j}(\boldsymbol{\hat{p}}_{ci,j+1} - \boldsymbol{\hat{p}}_{ci,j} - \mathcal{G}_{i,j}(s_i + j k + k) +\\
        & \phantom{1234561232314}  \mathcal{G}_{i,j}(s_{i} + j k))\| \le \phi_v h \quad \lor i \in \Lambda_p
    \end{aligned}
    \label{human_constraint}
\end{align}
where $\phi_x$ and $\phi_v$ are the thresholds for positions and velocities, respectively. $\boldsymbol{D}_{i,j}= \boldsymbol{I} - \mathcal{D}_{i,j}(s_i+ j k)\mathcal{D}_{i,j}(s_i + j k )^T$. Through the threshold, the sleeves can move within a range of deviation from the arm, which allow us to pull user's arm and adjust clothing deformation by contact, promoting the completion of the task. The above constraints prevent excessive stretching of clothing at contact points, replacing the safety loss $\mathcal{J}_{safe}$ and ensuring the force of clothing on users is within a safe threshold. For convenience, the above all constraints are integrated as $g_j(\boldsymbol{x}_{l,j}, \boldsymbol{v}_{l,j}) \le 0$.
\par
Based on the global trajectory, the local model is employed to provide higher frequency local compensation, which is conducted based on the following optimization problem:
\begin{align}
    \begin{split}
         \boldsymbol{v}_l = &\underset{\boldsymbol{v}_l}{\text{argmin}} \phantom{12} \mathcal{J}_{task}\\ 
         \text{subject to} \phantom{13} &\boldsymbol{x}_{l,j+1} = f_{loc}(\boldsymbol{x}_{l,j}, \boldsymbol{v}_{l,j}) \\
        & g_j(\boldsymbol{x}_{l,j}, \boldsymbol{v}_{l,j}) \le 0 \\
    \end{split}
    \label{loc_mpc}
\end{align}
where $f_{loc}$ is the local model defined in (\ref{loc-func1}), (\ref{loc-func2}). The final control law is derived as $\boldsymbol{\hat{v}}_r = \boldsymbol{v}_g + \boldsymbol{v}_l$.  Notably, before each round of local model solution, we truncate and extend the global solution based on the current time. The global trajectory is also divided into more segments for analysis with higher temporal resolution. Afterwards, we use $f_{sim}$ to calculate the global trajectory that matches new $\boldsymbol{v}_g$, as the basis for (\ref{loc_mpc}).

\subsection{Solution for Optimal Problem with Constraints}
To solve the optimization problem (\ref{loc_mpc}), we extend the constrained-Differential Dynamic Programming (CDDP) algorithm \cite{XieDifferentialDynamicProgramming2017} to deal with the fabric state constraints. 
\par
The standard DDP method utilizes a quadratic function to construct the optimal cost function from any time to the end (known as the cost-to-go function). In the backward pass, a lower-order approximated system function along the nominal trajectory is employed to update the cost-to-go function. In the forward pass, the nominal trajectory is re-calculated by solving the optimal input from the cost-to-go function. With the two processes conducted alternately, the nominal trajectory converges as the optimal trajectory. 
\par
In CDDP algorithm, an active set is introduced to represent critical inequality constraints. With the active constraint functions approximated by a first-order form, each step input solved in backward and forward pass is limited to meet the corresponding constraints. This method requires that the constraint dimension of each step is lower than the control dimension and is related to the control input, which ensures the existence of feasible control inputs that satisfy the constraints. In our problem, the constraints is introduced to describe the fabric behavior, parts of which are independent of the control inputs and may have a higher dimension in the local states, violating the conditions for CDDP.
\par
For convenience, denote the cost function at each step as $\mathcal{L}_j(\boldsymbol{x}_{l,j},\boldsymbol{v}_{l,j})$. $\mathcal{L}_j$ is quadratic function based on (\ref{Jpi}), (\ref{Jdi}), (\ref{Jwn}). Then, the cost-to-go function ${V}_j$ and the action value function $Q_j$ at each step satisfy:
\begin{align}
    \begin{split}
    &{V}_j(\boldsymbol{x}_{l,j}) = \min_{g_j(\boldsymbol{x}_{l,j},\boldsymbol{v}_{l,j})\le 0} {Q}_j(\boldsymbol{x}_{l,j}, \boldsymbol{v}_{l,j}) \\
    &{Q}_j(\boldsymbol{x}_{l,j}, \boldsymbol{v}_{l,j})= \mathcal{L}_j(\boldsymbol{x}_{l,j},\boldsymbol{v}_{l,j}) + V_{j+1}(f_{loc}(\boldsymbol{x}_{l,j}, \boldsymbol{v}_{l,j}))\\ 
    \end{split} 
    \label{VQ-define}
\end{align} 
As $V_j$, $\mathcal{L}_j$ are quadratic functions and $f_{loc}$ is a linear function, ${Q}_j$ is also quadratic. For convenience, we re-define $Q_j$ in terms of deviation from the nominal trajectory, which can be written as: 
\begin{align}
    \begin{split}
        &  Q_j(\boldsymbol{\delta}_{x,j}, \boldsymbol{\delta}_{v,j})= Q_{j}(\boldsymbol{0}, \boldsymbol{0}) + \boldsymbol{Q}^T_{x,j}\boldsymbol{\delta}_{x,j} + \boldsymbol{Q}^T_{v,j} \boldsymbol{\delta}_{v,j} +  \\
        &  \phantom{123458} \frac{1}{2}(\boldsymbol{\delta}_{x,j}^T\boldsymbol{Q}_{xx,j}\boldsymbol{\delta}_{x,j} + \boldsymbol{\delta}_{v,j}^T\boldsymbol{Q}_{vv,j}\boldsymbol{\delta}_{v,j})+\boldsymbol{\delta}_{v,j}^T\boldsymbol{Q}_{vx,j}\boldsymbol{\delta}_{x,j}
    \end{split}
\end{align}
The deviation $\boldsymbol{\delta}_{x,j},\boldsymbol{\delta}_{v,j}$ also satisfy the linear model, which can be denoted as:
\begin{align}
    \boldsymbol{\delta}_{x,j+1} = \boldsymbol{f}_x \boldsymbol{\delta}_{x,j} + \boldsymbol{f}_v \boldsymbol{\delta}_{v,j}
    \label{loc-model}
\end{align}
\par
Denote the subset of active constraints as $\hat{g}_j$. The linearization of the active constraints satisfy:
\begin{align}
    \boldsymbol{\hat{g}}_{x,j}\boldsymbol{\delta}_{x,j} + \boldsymbol{\hat{g}}_{v,j}\boldsymbol{\delta}_{v,j} = 0
    \label{current-constraint}
\end{align}
The constraints independent of $\boldsymbol{v}_l$ or with a higher dimension than $\boldsymbol{v}_l$ cause the null space of $\boldsymbol{\hat{g}}_{v,j}^T$ to be non-empty. Denote $\boldsymbol{\hat{g}}_{v,j}^+$ as a set of orthogonal bases in the null space of $\boldsymbol{\hat{g}}_{v,j}^T$. $\boldsymbol{\hat{g}}_{x,j}\boldsymbol{\delta}_{x,j}$ has no components in the null space, which can be written as:
\begin{align}
    (\boldsymbol{\hat{g}}_{v,j}^+)^T\boldsymbol{\hat{g}}_{x,j}\boldsymbol{\delta}_{x,j} = 0
    \label{dgxj}
\end{align}
By calculating the constraint (\ref{dgxj}) at the step $j+1$ and replacing $\boldsymbol{\delta}_{x,j+1}$ by (\ref{loc-model}), we obtain the equations to ensure the feasibility of future constraints:
\begin{align}
    (\boldsymbol{\hat{g}}_{v,j}^+)^T\boldsymbol{\hat{g}}_{x,j}\boldsymbol{f}_{x}\boldsymbol{\delta}_{x,j}+(\boldsymbol{\hat{g}}_{v,j}^+)^T\boldsymbol{\hat{g}}_{x,j}\boldsymbol{f}_{v}\boldsymbol{\delta}_{v,j}  = 0
    \label{future-constraint}
\end{align}
Notably, the additional constraints introduced in (\ref{dgxj}) can also create new null space. When transmitting the future constraints to the previous step, we use the combination of (\ref{current-constraint}) and (\ref{future-constraint}) to calculate the limitation on $\delta_{x,j}$, ensuring the feasibility of all constraints.
\par
Thus, in the backward pass, the function $V_j$ can be updated as:
\begin{align}
        \begin{split}
         &\min_{\boldsymbol{\delta}_{v,j}} \frac{1}{2}\boldsymbol{\delta}_{v,j}^T \boldsymbol{Q}_{vv,j}\boldsymbol{\delta}_{v,j} + \boldsymbol{\delta}_{v,j}^T\boldsymbol{Q}_{vx,j}\boldsymbol{\delta}_{x,j} + \boldsymbol{Q}_{v,j}^T\boldsymbol{\delta}_{v,j} \\
         &\phantom{1}\text{subject to} \phantom{13} \boldsymbol{\hat{g}}_{x,j}\boldsymbol{\delta}_{x,j} + \boldsymbol{\hat{g}}_{v,j}\boldsymbol{\delta}_{v,j} = 0 \\
         &\phantom{1\text{subject to} 13} \boldsymbol{c}_{j+1}\boldsymbol{f}_x\boldsymbol{\delta}_{x,j} + \boldsymbol{c}_{j+1}\boldsymbol{f}_v\boldsymbol{\delta}_{v,j} = 0 \\
    \end{split}
    \label{constraint-mpc}
\end{align}
where $\boldsymbol{c}_j = \left(\begin{bmatrix}
    \boldsymbol{\hat{g}}_{v,j} \\ \boldsymbol{c}_{j+1}\boldsymbol{f}_v
\end{bmatrix}^+\right)^T\begin{bmatrix}
    \boldsymbol{\hat{g}_{x,j}} \\  \boldsymbol{c}_{j+1}\boldsymbol{f}_x
\end{bmatrix}$.
\par
The solution for (\ref{constraint-mpc}) can be expressed analytically through KKT conditions, which is written as:
\begin{align}
    \begin{split}
    &\boldsymbol{Y}
    \begin{bmatrix}
    \boldsymbol{\delta}_{v,j} \\
    \boldsymbol{\lambda}_{1,j}    \\
    \boldsymbol{\lambda}_{2,j}
    \end{bmatrix}= -\begin{bmatrix}
        \boldsymbol{Q}_{vx,j} \\ \boldsymbol{\hat{g}}_{x,j} \\ \boldsymbol{c}_{j+1} \boldsymbol{f}_x 
    \end{bmatrix} \boldsymbol{\delta}_{x,j} -
    \begin{bmatrix}
        \boldsymbol{Q}_{v,j} \\ \boldsymbol{\sigma}_{1,j} \\ \boldsymbol{\sigma}_{2,j}
    \end{bmatrix}\\
    & \phantom{}\boldsymbol{Y} =\begin{bmatrix}
        \boldsymbol{Q}_{vv,j} & \boldsymbol{\hat{g}}^T_{v,j} & (\boldsymbol{c}_{j+1}\boldsymbol{f}_v)^T \\
        \boldsymbol{\hat{g}}_{v,j} & 0 & 0 \\
        \boldsymbol{c}_{j+1} \boldsymbol{f}_v & 0 & 0 
    \end{bmatrix}
    \end{split}
    \label{KKT-condition}
\end{align}
$\boldsymbol{\sigma}_{1,j} \ge 0$ and $\boldsymbol{\sigma}_{2,j} \ge 0$, representing the slack variables for inequality constraints. $\boldsymbol{\lambda}_{1,j}$, $\boldsymbol{\lambda}_{2,j}$ are the corresponding Lagrangian multipliers. According to the current active set, these constraints are all active with $\boldsymbol{\sigma}_{1,j} =0$ and $\boldsymbol{\sigma}_{2,j} = 0$. Then, by utilizing $\boldsymbol{\delta}_{u,j}$, $\boldsymbol{\lambda}_{1,j}$, $\boldsymbol{\lambda}_{2,j}$ as variables, we can obtain $\boldsymbol{\delta}_{v,j}$ as a function of $\boldsymbol{\delta}_{x,j}$:
\begin{align}
    \boldsymbol{\delta}_{v,j} = \boldsymbol{K}_j \boldsymbol{\delta}_{x,j}+ \boldsymbol{k}_j
    \label{optim-deltav}
\end{align}
where $\boldsymbol{K}_j$ and $\boldsymbol{k}_j$ are feedback gain and the open-loop term, with details provided in \cite{XieDifferentialDynamicProgramming2017}. By applying (\ref{optim-deltav}) to (\ref{VQ-define}), we can update the cost-to-go function iteratively backwards.
% \begin{align}
%     \begin{split}
%         &\boldsymbol{K}_{j} = -\boldsymbol{H}_j\boldsymbol{Q}_{vx,j} + \boldsymbol{W}_j^T \boldsymbol{D}^T_j\\
%         &\boldsymbol{k}_j = -\boldsymbol{H}_j\boldsymbol{Q}_{v,j} \\
%         &\boldsymbol{W}_j = (\boldsymbol{C}^T_j\boldsymbol{Q}_{vv,j}^{-1}\boldsymbol{C}_j)^{-1}\boldsymbol{C}_{j}^T \boldsymbol{Q}_{vv,j}^{-1}\\
%         &\boldsymbol{H}_j = \boldsymbol{Q}_{vv,j}^{-1}(\boldsymbol{I}- \boldsymbol{C}_j\boldsymbol{W}_j) \\
%         &\boldsymbol{C}_j = [\boldsymbol{\hat{g}}^T_{v,j} \quad (\boldsymbol{c}_{j+1}\boldsymbol{f}_v)^T ] \\
%         &\boldsymbol{D}_j = [\boldsymbol{\hat{g}}_{x,j}^T \quad (\boldsymbol{c}_{j+1}\boldsymbol{f}_v)^T]
%     \end{split}
% \end{align}
\par
In the forward pass, a new nominal trajectory is calculated by the control law (\ref{optim-deltav}). Meanwhile, we update the active set about the new nominal trajectory. To determine the influence of different constraints, we reduce the Lagrange multipliers to 0 and convert them into the corresponding slack variables $\boldsymbol{\sigma}_{1,j}$, $\boldsymbol{\sigma}_{2,j}$. In this case, $\boldsymbol{\delta}_{v,j}$ reaches the optimal solution without constraints at the $j$th step. $\boldsymbol{\sigma}_{1,j}$ and $\boldsymbol{\sigma}_{2,j}$ are solved as:
\begin{align}
    \begin{split}
         & \boldsymbol{\sigma}_{1,j} = -\boldsymbol{\hat{g}}_{x,j}\boldsymbol{\delta}_{x,j} + \boldsymbol{\hat{g}}_{v,j}\boldsymbol{Q}_{vv,j}^{-1}(\boldsymbol{Q}_{vx,j}\boldsymbol{\delta}_{x,j} + \boldsymbol{Q}_{v,j}) \\
         & \boldsymbol{\sigma}_{2,j} = -\boldsymbol{c}_{j+1}\boldsymbol{f}_x\boldsymbol{\delta}_{x,j} + \boldsymbol{c}_{j+1}\boldsymbol{f}_v\boldsymbol{Q}_{vv,j}^{-1}(\boldsymbol{Q}_{vx,j}\boldsymbol{\delta}_{x,j} + \boldsymbol{Q}_{v,j}) \\
    \end{split}
\end{align}
Denote the complete slack variable for constraints $\boldsymbol{\hat{g}}_j$ as $\boldsymbol{\sigma}_j$. $\boldsymbol{\sigma}_{1,j}$ and $\boldsymbol{\sigma}_{2,j}$ impact its variations outside and inside the null space of $\boldsymbol{\hat{g}}_{v,j}^T$ respectively. Through the $\boldsymbol{c}_{j+1}$ component in $\boldsymbol{c}_j$, $\boldsymbol{\sigma}_{2,j}$ can affect the null space from the $(j+2)$th step to the end. Thus, $\boldsymbol{\sigma}_j$ is calculated as:
\begin{align}
    \begin{bmatrix}
        \boldsymbol{\sigma}_j \\
        \boldsymbol{\sigma}_{n,j} \\ 
    \end{bmatrix} = \begin{bmatrix}
        \boldsymbol{\sigma}_{1,j} \\ \boldsymbol{0}
    \end{bmatrix} + 
    \begin{bmatrix}
        \boldsymbol{\hat{g}}_{v,j} \\ \boldsymbol{c}_{j}\boldsymbol{f}_v
    \end{bmatrix}^+ (\boldsymbol{\sigma}_{2,j-1} + \boldsymbol{\sigma}_{n,j-1})
    \label{sigma_j}
\end{align}
where $\boldsymbol{\sigma}_{n,j}$ describes the influence of $\boldsymbol{\sigma}_{2,j-1}$ on the $(j+1)$ step. $\boldsymbol{\sigma}_{n,0}$ is defined as $\boldsymbol{0}$. To update the active set, we check constraints not in $\boldsymbol{\hat{g}}_j$ and update $\sigma_j$ by (\ref{sigma_j}). Constraints reaching boundary conditions are added, while constraints with corresponding rows positive in $\sigma_j$ are removed.
\par
In practice, We introduce a line search strategy. When the optimization problem results in infeasible solutions, we shorten the update step to enhance the solution performance. To ensure that the new solution with line search satisfies the constraints, we replace $\boldsymbol{\delta}_{v,j}$ with $e\boldsymbol{\delta}_{v,j}$ in the object function of (\ref{constraint-mpc}), while keeping $\boldsymbol{\delta}_{v,j}$ unchanged in the constraints. $e$ is the line search factor. Based on the new optimization problem, $\boldsymbol{K}_j$, $\boldsymbol{k}_j$ with $e$ are obtained as the control law for updating the nominal trajectory in forward pass. Initially, $e=1$, which is reduced when the new nominal trajectory obeys the constraints or has a higher loss.
\par 
The whole process to solve (\ref{loc_mpc}) is shown as Algorithm \ref{cddp}.
\begin{algorithm}
    \SetAlgoLined
    \KwData{Maximum step $N_o$, termination indicator $\phi_t$} 
    \KwIn{Global trajectory $(\boldsymbol{x}_g, \boldsymbol{v}_g)$}
    initialize nominal local trajectory $(\boldsymbol{x}_l, \boldsymbol{v}_l)$ as zeros\;
    \For{$j=0, 1, \cdots, N-1$}{
        initialize active constraints $\boldsymbol{\hat{g}}_j$ by $\boldsymbol{x}_{l,j}$\;
    }
    $k \gets 0, d \gets  2 \cdot \phi_t$ \;
    $c_{now} \gets \mathcal{J}_{task}(\boldsymbol{x}_g + \boldsymbol{x}_l, \boldsymbol{v}_g + \boldsymbol{v}_l)$ \;
    \While{$k < N_o$ and $d > \phi_t$}{
        $e \gets 1, c_{new} \gets 2 \cdot c_{now}$ \;
        $V_N \gets 0$ \;
        \For{$j=N-1, N-2, \cdots, 0$}{
            update $Q_j$ by (\ref{VQ-define}) \;
            update $V_j$ by (\ref{VQ-define}) and (\ref{optim-deltav})\;
        }
        \While{$c_{new} > c_{now}$ and $k < N_o$}{
            $k \gets k+1$ \;
            $\boldsymbol{\delta}_{x,0} \gets \boldsymbol{0}, \boldsymbol{\sigma}_{n,0} \gets \boldsymbol{0}$\;
            \For{$j=0, 1, \cdots, N-1$}{
                solve $\boldsymbol{K}_j, \boldsymbol{k}_j$ with $e$ from (\ref{KKT-condition})\;
                $\boldsymbol{x}_{n,j} \gets \boldsymbol{x}_{l,j} + \boldsymbol{\delta}_{x,j}$\;
                $\boldsymbol{v}_{n,j} \gets \boldsymbol{v}_{l,j} + \boldsymbol{K}_j\boldsymbol{\delta}_{x,j} + \boldsymbol{k}_j$ \;
                $\boldsymbol{\delta}_{x,j+1} \gets f_{loc}(\boldsymbol{x}_{n,j}, \boldsymbol{v}_{n,j}) - \boldsymbol{x}_{l, j+1}$ \;
                update $\boldsymbol{\sigma}_j, \boldsymbol{\sigma}_{n,j}$ by (\ref{sigma_j}) \;
            }
            $c_{new} \gets \mathcal{J}_{task}(\boldsymbol{x}_g + \boldsymbol{x}_n, \boldsymbol{v}_g + \boldsymbol{v}_n)$ \;
            \eIf{$c_{new} > c_{now}$}{
                $e \gets e/2$\;
            }
            {
                $ d\gets c_{new} - c_{now}$ \;
                $c_{now} \gets c_{new}$ \;
                $(\boldsymbol{x}_l,\boldsymbol{v}_l) \gets (\boldsymbol{x}_n, \boldsymbol{v}_n)$ \;
                \For{$j=0, 1, \cdots, N-1$}{
                    update $\boldsymbol{\hat{g}}_j$ by $\boldsymbol{x}_{l,j}$ and $\boldsymbol{\sigma}_j$\;
                }
            }
        }
    }
    \Return $(\boldsymbol{x}_l, \boldsymbol{v}_l)$
\caption{The approach to solving (\ref{loc_mpc})}
\label{cddp}
\end{algorithm}

\section{Clothing Tracking and Human Prediction}
\label{sec:perception}
\subsection{Reconstruction and Tracking for Clothing}
To adopt diverse garments and provide feedback, we reconstruct garment meshes $\boldsymbol{p}_i$ from point cloud and extract the sleeve states $\boldsymbol{x}_{ci}$ for garment control.
\par
The GarmentNets framework \cite{ChiGarmentNetsCategoryLevel2021} serves as our foundation for reconstruction. Specifically, a canonical space with space resolution $c_x \times c_y \times c_z$ is defined for coats first. In this space, the clothing is fully extended in a standardized pose. Then the point cloud's probability distribution in this canonical space is computed to establish a mapping from partial observations to the canonical representation. This mapping enables analysis of the winding number field and the deformation flow from canonical to observed spaces. The smooth 3D garment structure is extracted through winding number field, with mesh sampling from the deformation flow yielding the current garment mesh. For implementation details, we refer readers to \cite{ChiGarmentNetsCategoryLevel2021}.
\par
In practical applications, GarmentNets calculate separable probability heatmaps for each spatial dimension in canonical space, which leads to the inconsistency of distributions in different dimensions. To mitigate this, we precompute the distance field $\boldsymbol{W}_c \in \mathbb{R}^{\times c_x \times c_y \times c_z}$ from canonical space to garment grid as weights to punish the deviation from the clothing grid due to the inconsistency. Denote the probability heatmap along respective axes as $\boldsymbol{p}_x\in \mathbb{R}^{c_x\times 1 \times 1}$, $\boldsymbol{P}_y\in \mathbb{R}^{1\times c_y\times 1}$, $\boldsymbol{P}_z\in \mathbb{R}^{1 \times 1 \times c_z}$. The loss function for observed-to-canonical space mapping is:
\begin{align}
    \begin{split}
    J_{cano} = -\frac{1}{c_x\cdot c_y\cdot c_z}&\sum( \boldsymbol{W}_c \odot \log (1 - \boldsymbol{P}_x \odot \boldsymbol{P}_y \odot \boldsymbol{P}_z) +  \\
     &\boldsymbol{W}_{gt} \odot \log (\boldsymbol{P}_x \odot \boldsymbol{P}_y \odot \boldsymbol{P}_z))
    \end{split}
\end{align}
where $\boldsymbol{A}\odot \boldsymbol{B}$ represents the element-wise multiplication of $\boldsymbol{A}$ and $\boldsymbol{B}$. $\boldsymbol{W}_{gt} \in \mathbb{R}_{c_x\times c_y\times c_z}$ is the one-hot encoding of the corresponding canonical space position.
\par
Upon the reconstructed mesh, we segment both sleeves using the canonical garment model and select six uniformly distributed keypoints at each sleeve's cross-section. The mean position of these keypoints is utilized as the armhole position $\boldsymbol{p}_{ci}$ while the smallest component in PCA decomposition is derived as the armhole orientation $\boldsymbol{v}_{ci}$.
\par
During operation, we acquire the mapping of the current observation in the canonical space and the corresponding deformation flow from the updated reconstruction. By comparing the garment mesh with the current observation in the canonical space, mesh nodes $\boldsymbol{p}_i$ existing matched observations are identified as visible surface points and subsequently updated via the computed deformation flow. The final corrected garment is obtained by the fusion of these observed nodes with model predictions, ensuring clothing tracking throughout the dressing procedure.
\par
For dataset, we leveraged the CLOTH3D \cite{BerticheCLOTH3DClothed3D2020} to obtain canonical garment models and compute the corresponding winding number fields. Within the Assistive Gym framework \cite{EricksonAssistiveGymA2020}, the dressing tasks were executed across diverse garment and human models, which yields 11000 samples of comprehensive point cloud and garment mesh under occlusion scenarios typical of dressing operations. To mitigate sim-to-real gap, the training pipeline only utilized point cloud inputs, avoiding reliance on synthetic rendering features that may not generalize to real-world scenarios.

\subsection{Human Perception and Prediction}
To enable personalized robotic dressing assistance, we achieve human perception and prediction to adopt the user motion under occlusion. Specifically, we extract human poses from observations and leverage historical observation sequences to generate future motion forecasts, which enhances optimal control and corrects the update of occluded human poses.
\par
At each step, a pretrained RTMPose model \cite{JiangRTMWRealTime2024} is utilized to extract 2D human keypoints, which are subsequently reconstructed into 3D poses using the depth information from the calibrated RGBD camera. The visible keypoints are matched against the prior motion prediction through a minimum joint deviation criterion, yielding the new joint states to calculate the occluded points. For human motion prediction, we categorize human intent into passive and active behaviors. In passive scenarios, human adjusts their poses following garment motion to minimize garment contact forces. In active cases, human exhibits independent intent by moving toward arbitrary spatial targets. The former characterizes the human's passive adaptation to garment motion, while the latter models the spatio-temporal relationship of human motion intention during the dressing process. We employ an attention-based discriminator network to select between these prediction modalities. Then, a diffusion model \cite{MaoLeapfrogDiffusionModel2023} predicts arm trajectories conditioned on 20 steps historical motion sequences and the corresponding sleeve states. 
\par
As to the dataset, we employed the local garment model with relaxed human-garment contact constraints (\ref{human_constraint}) to solve clothing motion under passive human intention. In this case, the garment model only consider the deformation limits. The clothing moved without the influence of the human pose as the user collaborates perfectly. Then, by using the garment as arm constraints, we derived the minimal human actions, representing the passive dressing behavior. For active data, we randomly sampled target positions and generate arm trajectories toward them. The corresponding sleeve states were solved by the local model with the whole contact constraints. After random initialization of garment and human parameters, we collected 2,000 dressing sequences respectively for both scenarios.

\section{Experiments}
\subsection{Experimental Setup}
The experimental setup is illustrated in the upper left corner of Fig. \ref{pipeline}. The system comprises a dual-arm configuration employing two 7-degree-of-freedom Franka robotic manipulators. Each manipulator grasped the coat shoulder from the inner side to facilitate garment unfolding, thereby enabling subsequent manipulation tasks. Two Gemini 335 depth cameras were positioned on the anterior and posterior sides of the robotic arms to provide garment reconstruction and human pose estimation. Our algorithm was executed on a workstation equipped with an Intel(R) Core i9-13900HX processor, 31 GiB of RAM, and an NVIDIA RTX 4060 Desktop GPU, delivering updates to the local optimization problem at a frequency of $10Hz$.  
\par
Based on the perception results, we employed vertex clustering to simplify the garment mesh with 5 cm grid resolution, which ensures computational efficiency while maintaining collision detection accuracy. For human body representation, the user's morphology was reconstructed as a kinematic composition of cylindrical and spherical primitives, with joint positions dynamically updated through keypoint observations. The simulation scenario for the dressing task is shown in the upper right corner of Fig. \ref{pipeline}.
\par
To prevent inter-arm collisions during the dressing process, redundancy freedom was employed to maintain separation between the robotic manipulators. Let $\boldsymbol{q}_l\in\mathbb{R}^7$ and $\boldsymbol{q}_r\in\mathbb{R}^7$ denote the joint angles of the left and right manipulators, respectively. The desired end-effector positions are represented as $\hat{\boldsymbol{p}}_{rl}\in\mathbb{R}^3$ and $\hat{\boldsymbol{p}}_{rr}\in\mathbb{R}^3$, with corresponding end-effector Jacobian matrices $\boldsymbol{J}_l\in\mathbb{R}^{3\times 7}$ and $\boldsymbol{J}_r\in\mathbb{R}^{3\times 7}$. Additionally, denote the Jacobian matrices for the third link positions as $\boldsymbol{J}_{3l}\in\mathbb{R}^{3\times 7}$ and $\boldsymbol{J}_{3r}\in\mathbb{R}^{3\times 7}$. The control law for the manipulators is formulated as follows:  
\begin{align}
    \begin{split}
        & \boldsymbol{\dot{q}_l} = \boldsymbol{J}_l^{\dagger}(\boldsymbol{\hat{p}}_{rl} - \boldsymbol{p}_{rl})+ (\boldsymbol{I} - \boldsymbol{J}_l^{\dagger}\boldsymbol{J}_l)\boldsymbol{J}_{3l}^{\dagger}({\boldsymbol{\hat{p}}}_{rr} - \boldsymbol{p}_{rl})  \\
        & \boldsymbol{\dot{q}_r} = \boldsymbol{J}_r^{\dagger}(\boldsymbol{\hat{p}}_{rr} - \boldsymbol{p}_{rr}) + (\boldsymbol{I} - \boldsymbol{J}_r^{\dagger}\boldsymbol{J}_r)\boldsymbol{J}_{3r}^{\dagger}({\boldsymbol{\hat{p}}}_{rl} - \boldsymbol{p}_{rr})
    \end{split}
\end{align}
Given that the grasping points of the robotic arms were positioned on either side of the user during the dressing task, and the manipulator bodies are actively displaced outward, potential collisions between the dual-arm system and the human subject can be effectively mitigated.
\par

\subsection{Validation for Clothing Simulation}  
To validate the efficacy of our clothing  simulation, we conducted comparative experiments involving draping and placement operations using a coat model, benchmarking against the standard PD algorithm to quantitatively evaluate the computational efficiency and accuracy of our proposed method. The employed coat model comprises a 6,121-node mesh, whose geometric structure is similar to the garments used in the assisted dressing task. In the subsequent simulation, the garment was initialized in a normalized object state, with grasping points defined as the 10 proximal nodes to each shoulder position. Spring connections are established between these points and $\boldsymbol{p}_{rl}, \boldsymbol{p}_{rr}$, which represents the robot's impact on clothing.
\par
\begin{figure*}[htbp]\centering 
        \includegraphics[width=0.95\textwidth,trim=1 1 1 1,clip]{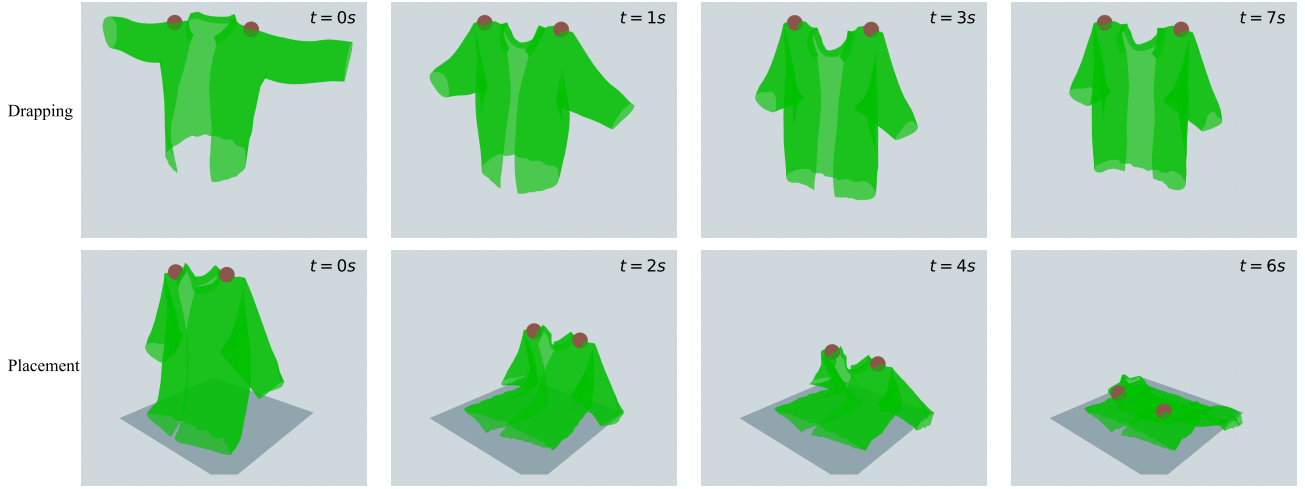}
        \label{sim-exp-result}
        \caption{Simulation results of our proposed algorithm with step $h = 0.01 s$ for both garment draping and placement tasks.}
\end{figure*}
\begin{figure}[!t]\centering
    \includegraphics[width=0.45\textwidth,trim=0 0 0 0,clip]{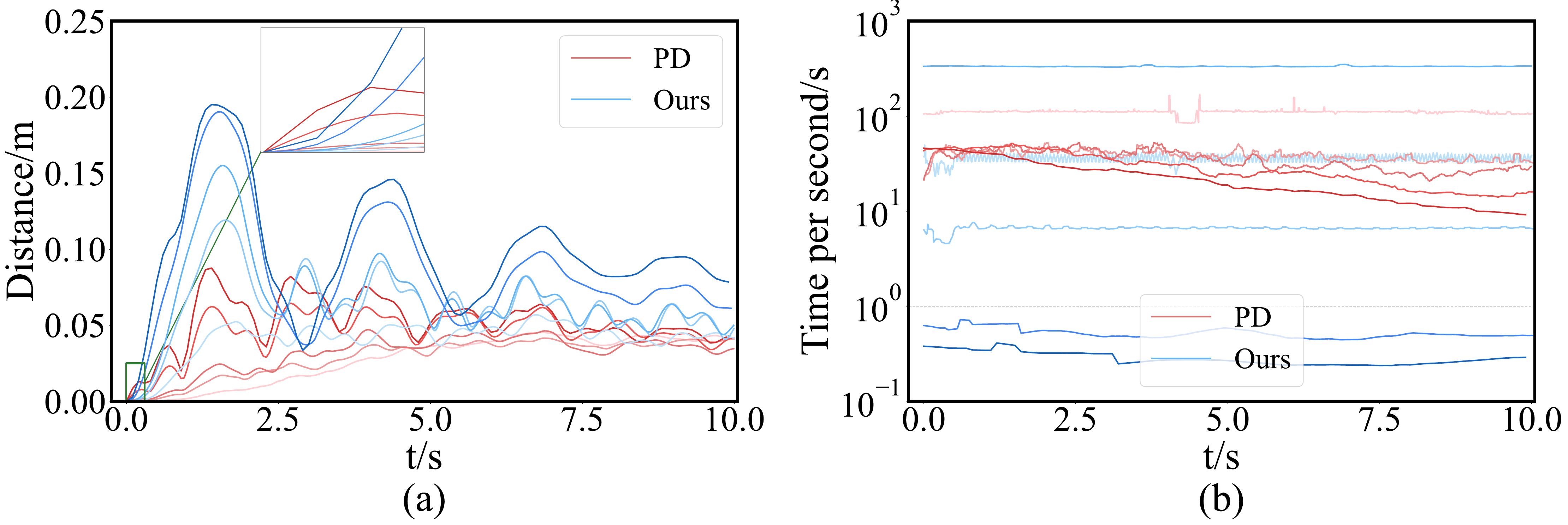}
        \label{sim-exp-1}
    \caption{The results of draping task. Our method and the PD algorithm are shown by blue and red, respectively. The time steps $h \in \{0.001s, 0.05s, 0.01s, 0.05s, 0.1s\}$ are represented by color gradient from light to dark. (a) The nodal mean distance with the high temporal resolutions PD simulation. (b) The computational cost for $1s$ of simulation.}
\end{figure}

In the draping simulation, the clothing was released from the normalized object state and stretched downward under the influence of gravity. The deformation process is illustrated in the first column of Fig. \ref{sim-exp-result}, which is calculated by our method with the step $h = 0.01s$. To evaluate the performance of our method under large time-step conditions, we conducted a systematic comparative analysis between our method and the PD algorithm across varying temporal resolutions (as $h \in \{0.001s, 0.005s, 0.01s, 0.05s, 0.1s\}$). The obtained results were compared against the high temporal resolutions PD simulation at a time step of $0.0001s$, thereby elucidating the different impacts of increased steps on both our approach and the PD method. The difference with the fine-grained simulation are presented in Fig. \ref{sim-exp-1} (a), while the computational cost required for $1s$ of simulation time is shown in Fig. \ref{sim-exp-1} (b). This metric measures the real-time performance of the method. A value below $1s$ indicates that real-time simulation can be achieved. The mean computational time per step for each configuration is tabulated in Table \ref{per-frame-time}. 
\par
As illustrated in Fig. \ref{sim-exp-result}, our method produces visually plausible and physically reasonable deformation predictions for garments in collision-free scenarios. Owing to its second-order accuracy guarantees, our approach derived the results more similar to high-resolution simulation  during the initial simulation steps, as highlighted by the green bounding box in Fig. \ref{sim-exp-1} (a). However, as simulation progresses, accumulation of numerical errors leads to gradually increasing deviation across all tested time steps. The distance peaks at approximately $t = 1.5 s$, which stems from the stabilization effect induced as the high-order terms in (\ref{standard-iteration}). This artificial damping accelerates energy dissipation within the system. In contrast, the PD method maintains better energy preservation at small step sizes, resulting in persistent fabric oscillations that diverge from our predictions. Over extended durations, the PD simulation eventually exhibits gradual energy decay and converges to a stable deformation state. At this time, our method achieves comparable accuracy to PD at sufficiently small step sizes. This performance characteristic represents our method's numerical damping properties, which aligns with real-world scenarios where natural energy dissipation occurs through air resistance and other damping effects.
\par
As to the simulation cost shown in Table \ref{per-frame-time}, PD method needs $0.110s$ for per step. However, due to growing nonlinearity in its implicit formulation, the solution cost increases synchronously with the step size. Despite employing larger time steps, the PD method still suffers from comparable computational overhead and increased simulation errors, resulting in poor performance. Instead, our approach maintains consistent computational efficiency across different step sizes $h$, demonstrating the superior real-time capability.
\par
\begin{figure}[!t]\centering
    \includegraphics[width=0.45\textwidth,trim=0 0 0 0,clip]{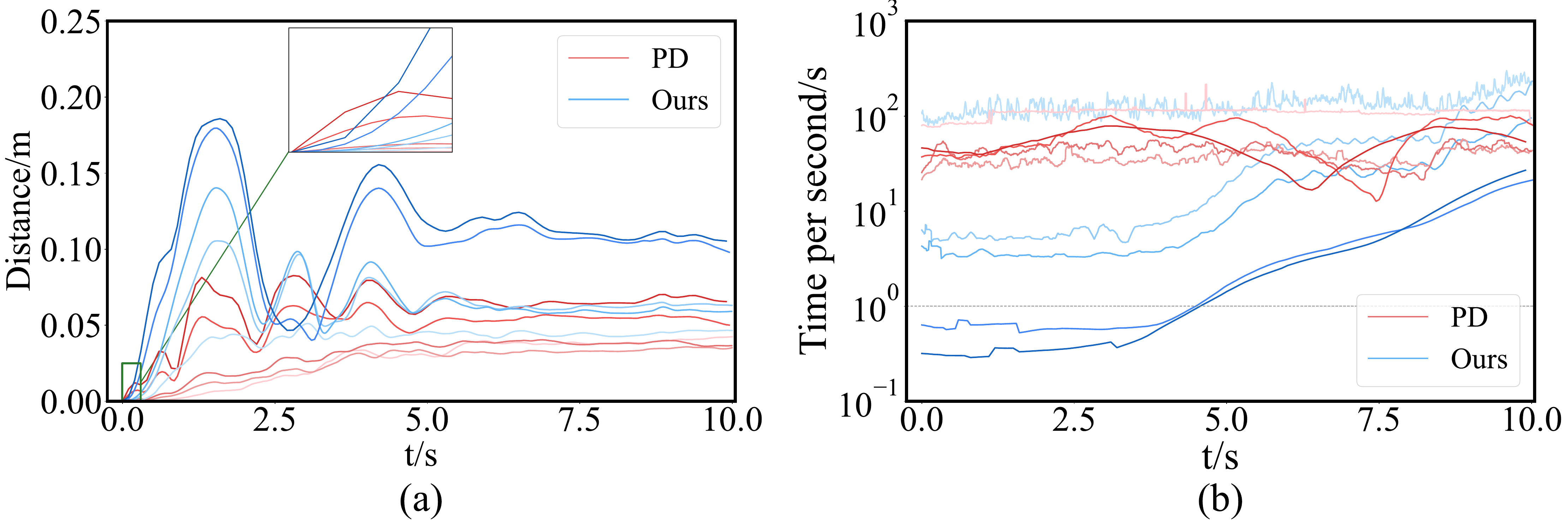}
        \label{sim-exp-2}
    \caption{The results of placement task. Our method and the PD algorithm are shown by blue and red, respectively. The time steps $h \in \{0.001s, 0.05s, 0.01s, 0.05s, 0.1s\}$ are represented by color gradient from light to dark. (a) The nodal mean distance with the high temporal resolutions PD simulation. (b) The computational cost for $1s$ of simulation.}
\end{figure}

To validate the simulation with contacts, we conducted experiments by placing and folding a coat on the horizontal table. During this process, the contact points between the garment and the surface gradually increased until full contact was achieved. This procedure is illustrated in the second column of Fig. \ref{sim-exp-result}. Similarly, we evaluated the computational efficiency and accuracy of our method versus PD method at different time steps ($0.001 s, 0.005 s, 0.01 s, 0.05 s$, and $0.1 s$). The results are presented in Fig. \ref{sim-exp-2}, and Table \ref{per-frame-time}. 
\par
Similar to the draping task, our method initially exhibited deviation during the early simulation phase. With the motion of grasping points, garment-table contact was established, and the method achieved comparable results to PD at small step sizes. Notably, the computational overhead of our method increased with growing contact, as (\ref{obtain-rf}) requires inverse analysis of forward process for all contact nodes. When contact involved all nodes, this incurred cubic  time complexity. In assisted dressing tasks, garment-body contact primarily occurs through sleeves, involving only a small subset of garment nodes. Besides, by reducing the mesh dimensionality, the number of contact points was further diminished. Given the $\mathcal{O}(n^3)$ complexity of matrix inversion, the contact prediction never dominates the overall computational cost.
\begin{table}[!t]
    \caption{Per-frame computational time for different simulation tasks.}
    \label{per-frame-time}
    \centering
    \begin{tabular}{l|c|c|c|c|c}
       \hline
    \diagbox{Tasks}{Step} &  $0.001s$ & $0.005s$  & $0.01s$ & $0.05s$ & $0.1s$    \\
      \hline
    PD in draping & $0.110s$ & $0.194s$ & $0.348s$ & $1.394s$ & $2.321s$ \\
      \hline
    Ours in draping& $0.036s$ & $0.033s$ & $0.034s$ & $0.026s$ & $0.029s$ \\
      \hline
    PD in placement & $0.110s$ & $0.184s$ & $0.421s$ & $2.938s$ & $5.249s$ \\
      \hline
    Ours in placement& $0.135s$ & $0.215s$ & $0.198s$ & $0.319s$ & $0.620s$ \\
      \hline
    \end{tabular}
\end{table}
\par
In summary, while our method trades marginal accuracy in highly dynamic scenarios, it can achieve plausible predictions while maintaining real-time performance. Given the forward and backward equations (\ref{force-contact}) and (\ref{backward-update}),  the computational bottleneck lies in the multiplication with $\boldsymbol{P}^{-1}$ and the calculation of $\boldsymbol{P}_{ic}^{-1}$. Denote the number of contact nodes as $n_c$. The time complexity of one whole step is $\mathcal{O}(n_c^3+n^2)$. In subsequent experiments, a refined mesh spacing of 0.05 cm (yielding approximately 1,800 nodes) was employed, enabling faster-than-real-time simulation at the step size $h = 0.01s$. Additionally, to ensure efficient global controller updates, simulation with $h =1s$ was employed for predictive optimization, while the initial values for local optimizer was calculated with $h = 0.01s$.
\par
Subsequently, we evaluated the differentiable ability of our simulation on a simplified single-sleeve dressing task. In this scenario, the garment was grasped at bilateral shoulder positions $\boldsymbol{p}_{rl}$, $\boldsymbol{p}_{rr}$, with six nodes sampled around the left armhole to compute the sleeve state $x_{cl}$. The human's left arm was raised, and the target position $x_{tar}$ was sequentially selected along the arm coordinate system. With $x_{tar}$ progressively moving along the arm, we maneuvered the grasping points to track $x_{tar}$ by the sleeve state $x_{cl}$, thereby accomplishing the garment dressing task. Based on the invertible simulation derivatives, we implemented the following control law:
\begin{align}
    \begin{split}
        &\boldsymbol{\dot{p}}_{rl} = \boldsymbol{K} (\frac{\partial \boldsymbol{x}_{cl}}{\partial \boldsymbol{x}_{n+1}}\frac{\partial \boldsymbol{x}_{n+1}}{\partial \boldsymbol{x}_{n}}\frac{\partial \boldsymbol{x}_n}{\partial \boldsymbol{p}_{rl}})^T (\boldsymbol{x}_{tar} - \boldsymbol{x}_{cl})\\
        &\boldsymbol{\dot{p}}_{rr} = \boldsymbol{K} (\frac{\partial \boldsymbol{x}_{cl}}{\partial \boldsymbol{x}_{n+1}}\frac{\partial \boldsymbol{x}_{n+1}}{\partial \boldsymbol{x}_{n}}\frac{\partial \boldsymbol{x}_n}{\partial \boldsymbol{p}_{rr}})^T (\boldsymbol{x}_{tar} - \boldsymbol{x}_{cl})\\
    \end{split}
\end{align}
where $\boldsymbol{K}$ is a proportional coefficient. 
\begin{figure}[!t]\centering
    \includegraphics[width=0.45\textwidth,trim=0 0 0 0,clip]{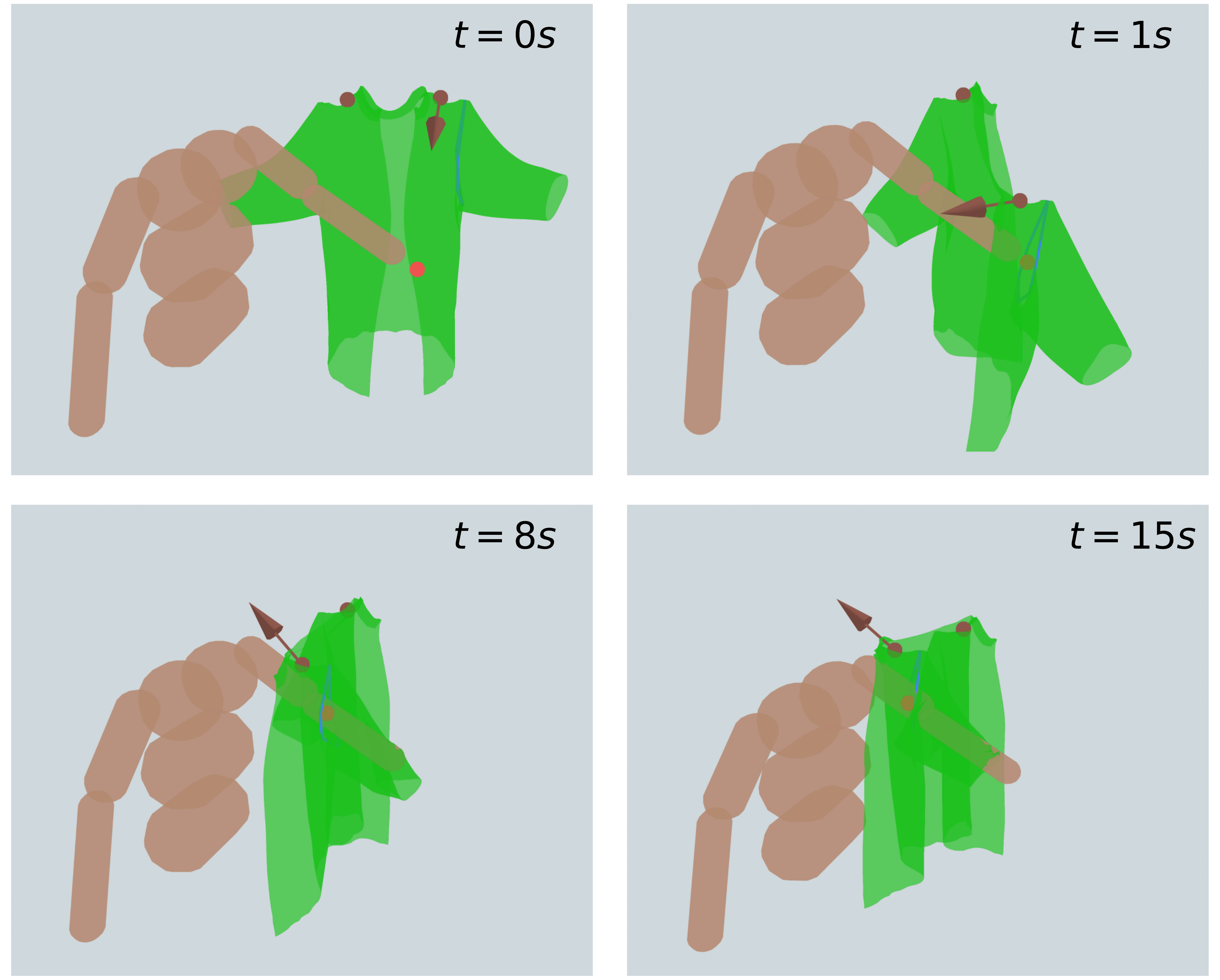}
    \caption{The results of single-arm dressing using the differentiable garment simulation. The target position $x_{tar}$ is represented by red points, while the grasping points are indicated by brown points. The target position $x_{tar}$ is represented by red points. The contour of armhole is shown in blue.}
    \label{sim-contr-result}
\end{figure}
\par
\begin{figure}[!t]\centering
    \includegraphics[width=0.45\textwidth,trim=10 0 0 0,clip]{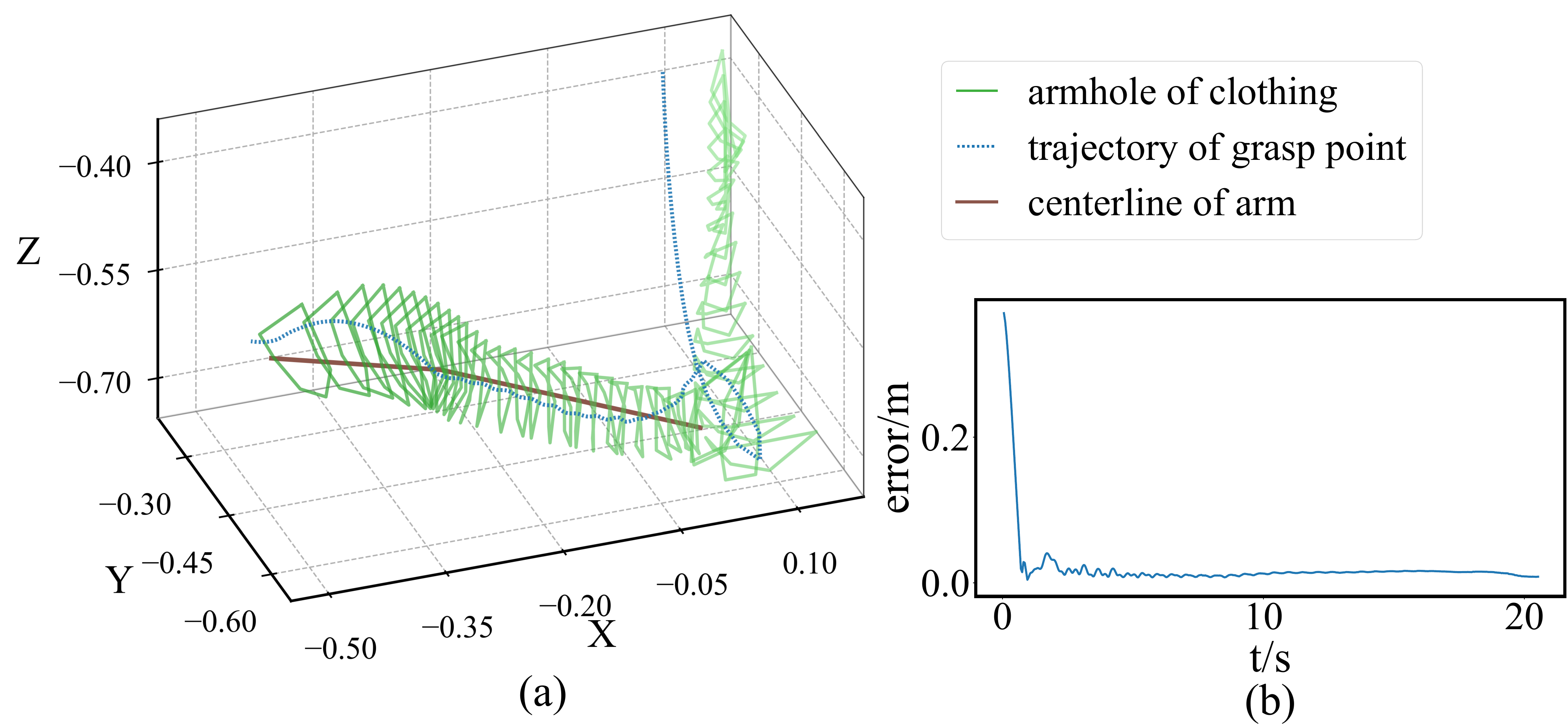}
    \caption{(a) The motion trajectories of the sleeve state $x_{cl}$ and grasp point $\boldsymbol{p}_{rl}$ during dressing. (b) The errors between the sleeve state $x_{cl}$ and target position $x_{tar}$.}
    \label{sim-contr-err}
\end{figure}

The dressing process is illustrated in Fig. \ref{sim-contr-result}. Since the left sleeve was closer to the left shoulder grasping point, $x_{cl}$ was predominantly influenced by this point, driven by the motion of $p_{rl}$. The corresponding manipulation trajectory is depicted in Fig. \ref{sim-contr-err} (a).
During the initial phase of the task, the target point $\boldsymbol{x}_{tar}$ was positioned at the  end of the arm, far from the sleeve state $x_{cl}$, resulting in large control input that drove rapid convergence to the target position. Subsequently, as $x_{tar}$ gradually translated along the arm, our controller can track $x_{tar}$ from hand to shoulder, maintaining minimal position error as shown in Fig. \ref{sim-contr-err} (b).  As $x_{cl}$ reached the shoulder position, the entire sleeve was successfully donned onto the arm, completing the single-arm dressing task. The experimental results validate the efficacy of our garment differentiable simulation method. 
\par
\subsection{Robot-Assisted Dressing Experiments} % 3
\begin{figure}[!t]\centering
    \includegraphics[width=0.3\textwidth,trim=0 0 0 0,clip]{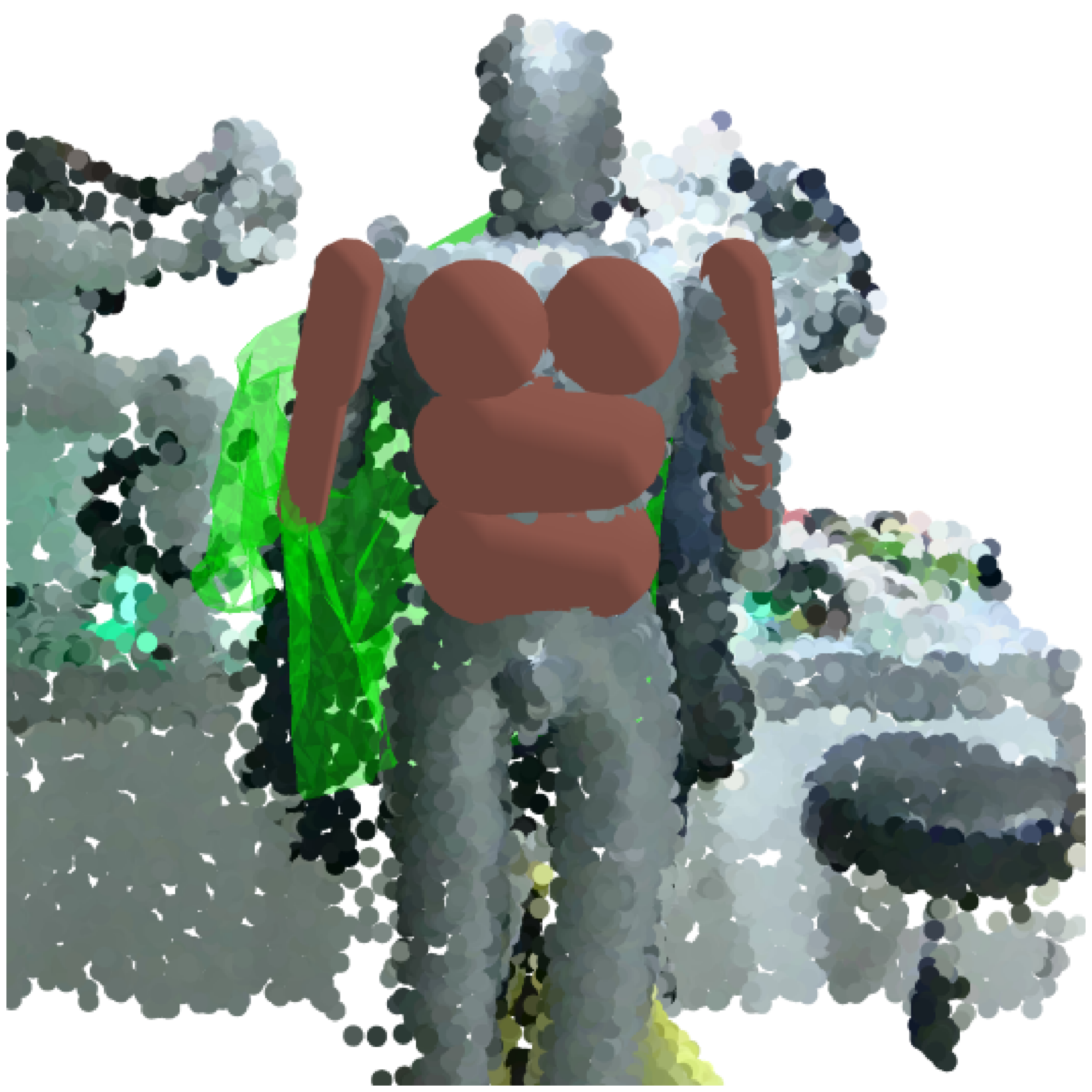}
    \caption{The reconstruction of the human body and garment in the point cloud.}
        \label{pc-scene}
\end{figure}

\begin{figure*}[htbp]\centering 
        \includegraphics[width=0.95\textwidth,trim=1 1 1 1,clip]{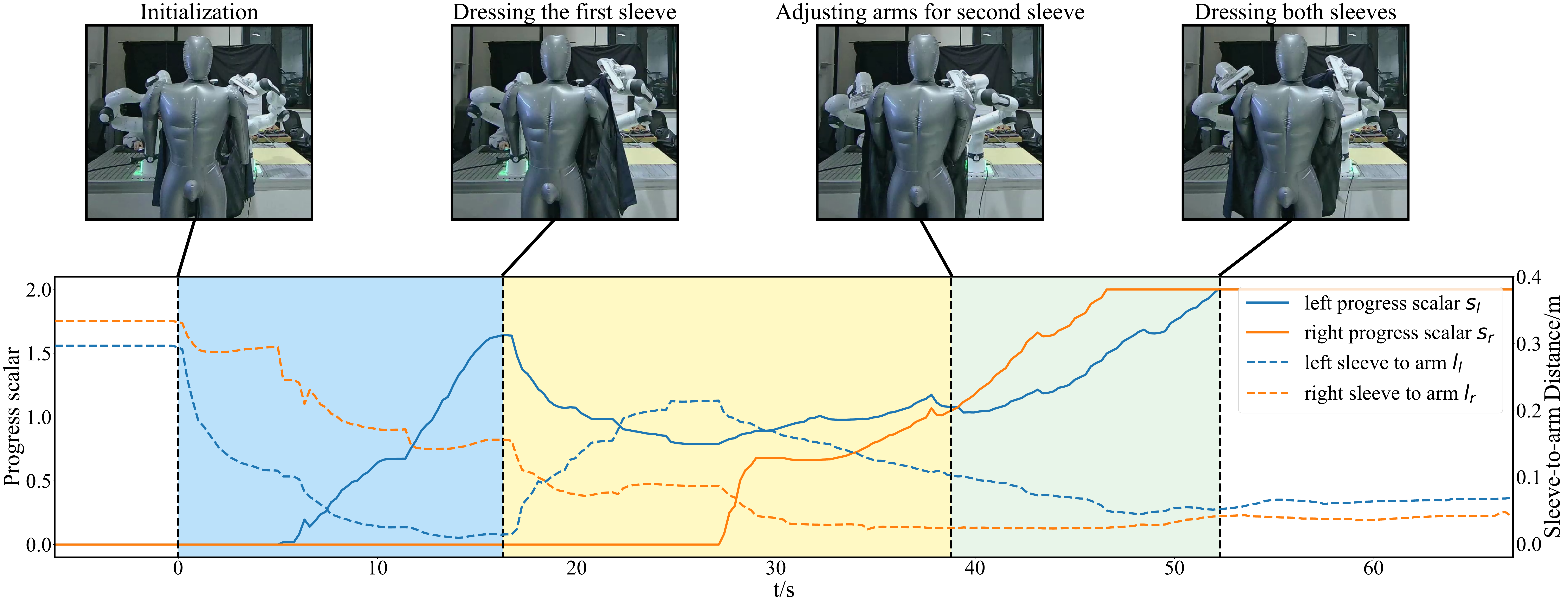}
        \caption{The task phase segmentation and the corresponding sleeve states during dummy dressing.}
        \label{task-segmentation}
\end{figure*}

To validate the efficacy of the assisted dressing methodology, we initially conducted experiments on a dummy. Due to the disparity between the human and dummy characteristics, the pose estimation derived from RTMPose exhibits inaccuracies. For convenience, we manually annotated keypoint positions and employed Cotracker\cite{KaraevCoTrackerItIs2024} for real-time tracking as a surrogate for human perception. Subsequently, a human body model was generated from these keypoints, while the garment meshes were reconstructed using the point cloud on the back. As shown in Fig. \ref{pc-scene}, the reconstructed human body and garments aligned with their actual positions in distance and scales. Given the absence of intent in dummy, we consistently utilized the passive intent for the behavior prediction, representing the dummy's compliant deformation to follow the garment manipulation.  The experimental results are presented in Fig. \ref{task-segmentation}.
\par
In Fig. \ref{task-segmentation}, we illustrate the progress scalar $s_i$ and the sleeve-to-arm distance $l_i$ for two sleeves in their corresponding arm coordinate systems. These variables quantitatively assessed the dressing task progression and bilateral sleeves dressing quality, respectively. Based on their temporal trends, the entire dressing procedure was naturally divided into three distinct phases: dressing the first sleeve, adjusting garment-body pose, and dressing the both sleeves simultaneously. 
\par
Initially, the garment exhibited global motion toward the left arm with the corresponding distance $l_l$ decreasing. After the left arm squeezed into the armhole, the sleeve was raised along the left arm with progress scalar $s_l$ increasing. Upon exceeding a predetermined threshold (0.5) for $s_l$, the system transitioned to adjusting body pose, redirecting the garment toward the second arm. In this case, the contact force with the left sleeve facilitated pose adaptation, bringing both arms into proximity for simultaneous sleeve donning. Finally, with the second arm inserted into the sleeve, both sleeves ascended along their respective arms until their progression scalars reach $s_i=2$, indicating task completion. Stage-wise snapshots in Fig. \ref{task-segmentation} demonstrates that our method has successfully provided dressing assistance for the dummy.
\par
\begin{figure}[!t]\centering
    \includegraphics[width=0.45\textwidth,trim=0 0 0 0,clip]{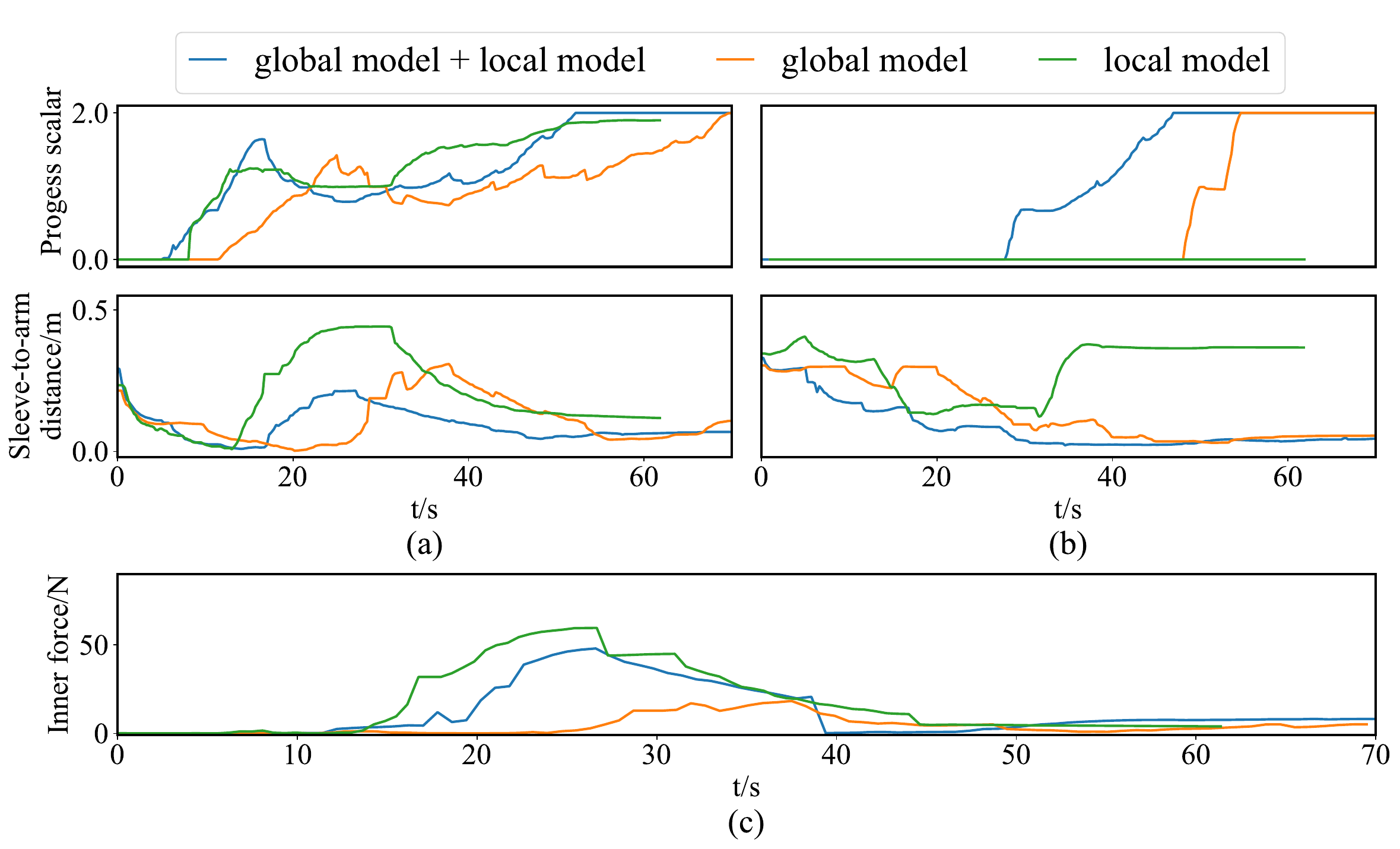}
    \caption{Evolution of the sleeve state and maximum simulated inner forces under different methods. (a) The state of the left sleeve in the arm coordinate system.  (b) The state of the right sleeve state in the arm coordinate system. (c) The maximum inner force caused by constrained deformation.}
        \label{faker-contr}
\end{figure}

To evaluate the contribution of each component in our approach, we conducted ablation studies examining different method combinations for robotic dressing on a dummy. Beyond the complete method (global+local method), we investigated two reduced variants: using only the global controller (\ref{global_mpc}) for control (denoted as global-only method), and employing solely the local linear model (\ref{loc-func2}) for updating the sleeve state $x_{ci}$ (denoted as local-only method). The resulting sleeve state transitions and maximum simulated inner forces across methods are presented in Fig. \ref{faker-contr}
% , with corresponding robotic manipulation trajectories depicted in Fig. \ref{faker-trajectory}.
% \begin{figure}[!t]\centering
%     \includegraphics[width=0.45\textwidth,trim=0 0 0 0,clip]{image/faker_trajectory.pdf}
%     \caption{Robotic motion trajectories for different method configurations.}
%         \label{faker-trajectory}
% \end{figure}
\par
In Fig. \ref{faker-contr} (a) and (b), we calculated the sleeve states within the arm coordinate system through cloth simulation. Through the incorporation of the local model in control, the complete method represented higher dressing efficiency than the global-only approach. The local compensation enhances the controller update frequency and enables dynamic response to real-time garment changes, accelerating state convergence toward target configuration. Conversely, due to the modeling inaccuracies, the local-only method exhibited larger $l_i$ when donning the second sleeve. Given that the local model (\ref{loc-func2}) is incapable of capturing sleeve opening orientation changes, it attempted to insert the second sleeve from inappropriate directions, which substantially increases dressing failure rates. 
% This results in the further outward grasping trajectory in Fig. \ref{faker-trajectory}. 
On the other hand, as shown in Fig. \ref{faker-contr} (c), the global model inclusive approaches incorporate force cost (\ref{Jsafe}) into control, thereby reducing peak deformation forces during dressing. In summary, for dummy dressing, the local-only method proves inadequate for donning the second sleeve due to its lacking deformation modeling capacity. While the global-only model alone achieves comparable results to the complete method for stationary targets, the complete method's higher control frequency and dressing efficiency become particularly advantageous for human dressing scenarios, as it can better accommodate subject's motion.
\begin{figure*}[htbp]\centering 
        \includegraphics[width=0.95\textwidth,trim=1 1 1 1,clip]{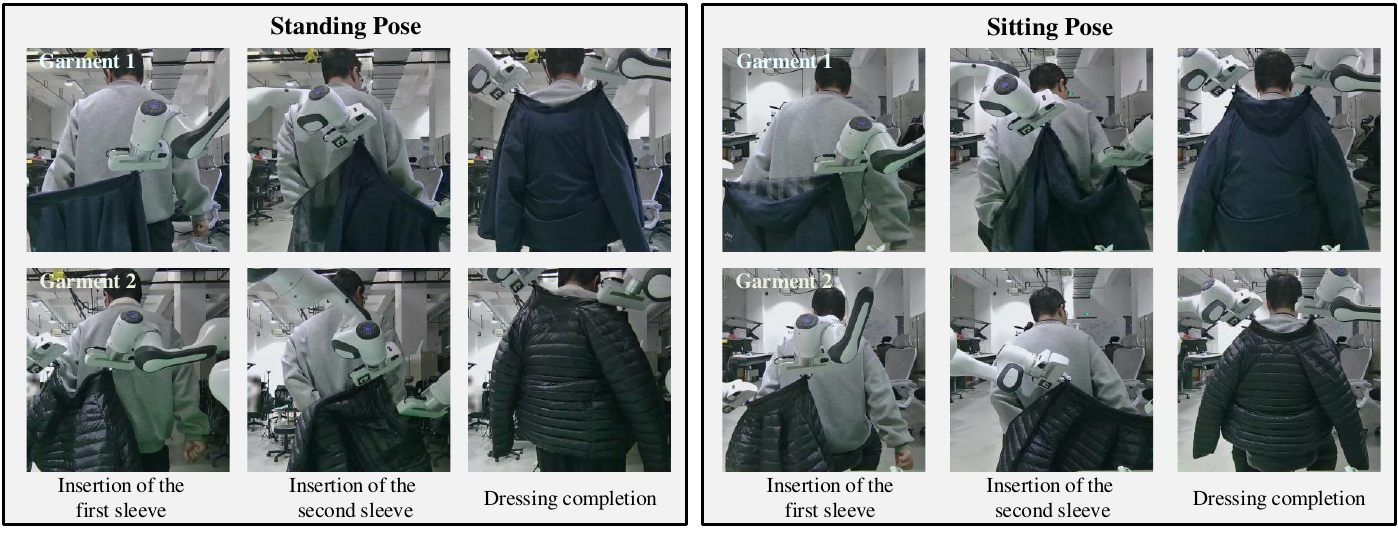}
        \caption{The assisted dressing outcomes for two garments in both standing and sitting poses with human passive motion. Each case is illustrated through three critical phases: the insertion of the first and the second sleeve, and final dressing completion.}
        \label{human-scene}
\end{figure*}
\begin{figure*}[htbp]\centering 
        \includegraphics[width=0.95\textwidth,trim=1 1 1 1,clip]{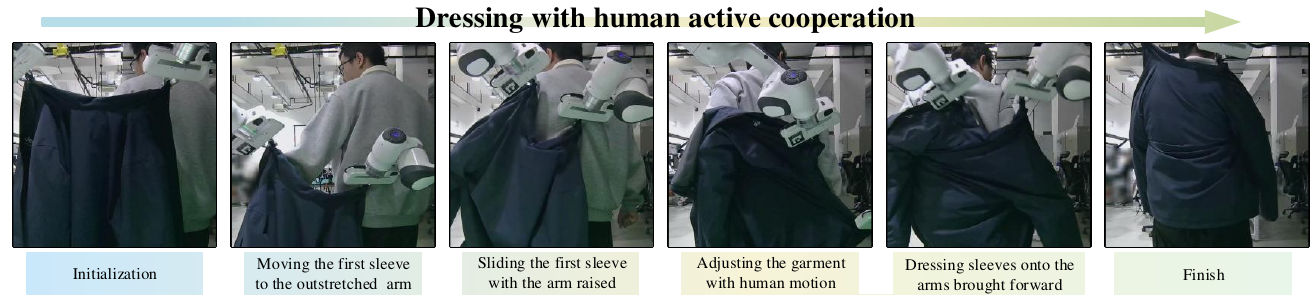}
        \caption{The assisted dressing process with active human cooperation. According to the variations in human strategy, the donning process is illustrated through two phases: the preparation for the sleeve insertion and upward sliding motion for each sleeve.
        }
        \label{human-scene_act}
\end{figure*}
\begin{figure}[!t]\centering
    \includegraphics[width=0.45\textwidth,trim=0 0 0 0,clip]{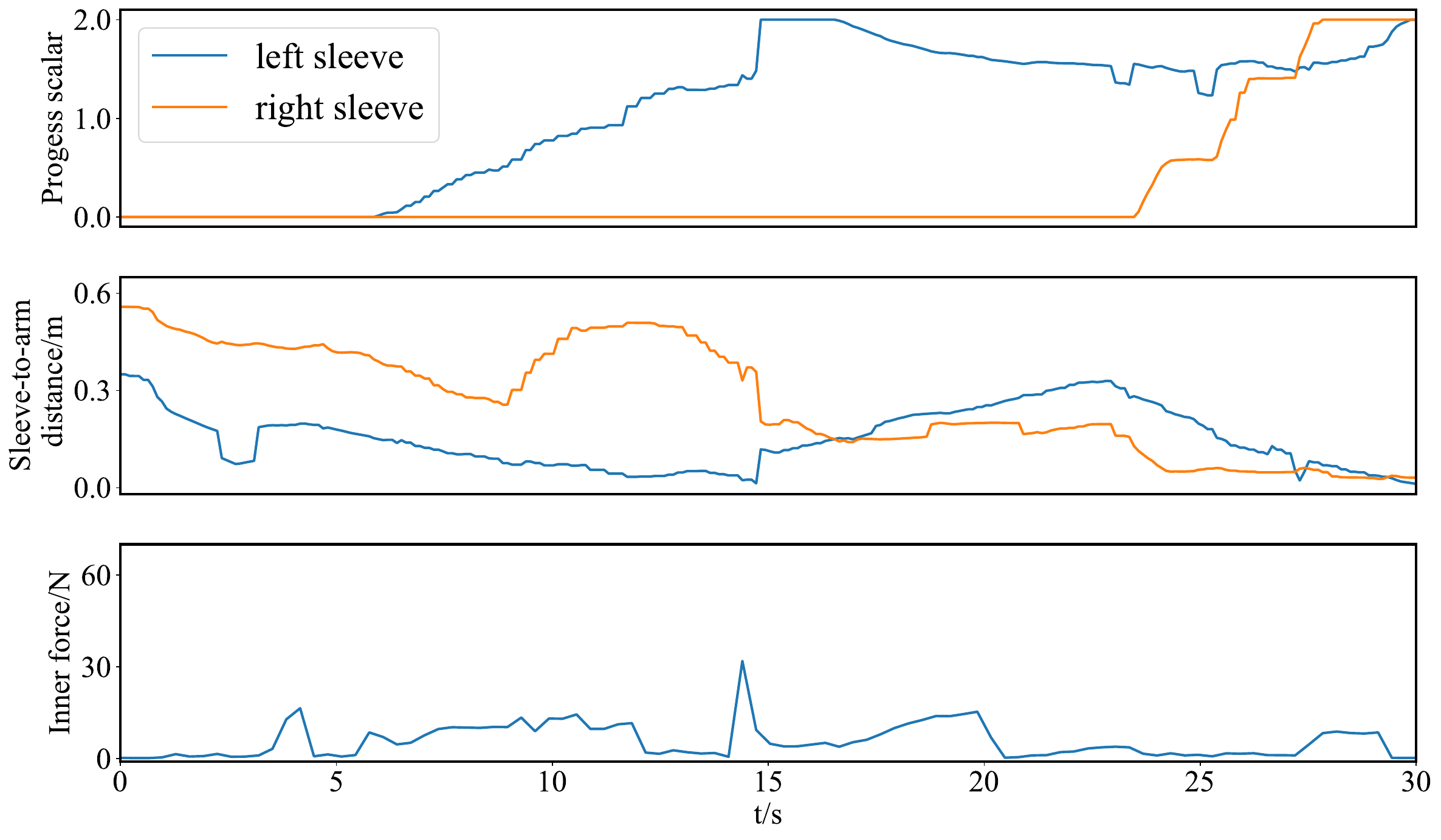}
    \caption{Evolution of the sleeve state and maximum simulated inner forces during active human cooperation.}
        \label{predict-err}
\end{figure}
\par
To validate the performance of our method in real-world dressing scenarios, we conducted comprehensive experiments with diverse garments, motion patterns, and human poses, as illustrated in Fig. \ref{human-scene}, \ref{human-scene_act}. Fig. \ref{human-scene} demonstrates the human passive dressing outcomes for two garments in both standing and sitting poses. In these trails, the participant compliantly followed garment guidance to adjust arm poses, enabling successful task completion. Experimental results demonstrate the robot's capability to properly position garments relative to arm locations, successfully don the first sleeve, and subsequently align the second sleeve while maintaining the first sleeve's position to complete the outerwear dressing procedure.
\par
We further investigated the dressing performance with active human cooperation, where the subject deliberately adjusted stance and actively extended arms toward the garment. The donning process for both sleeves under this condition is depicted in Fig. \ref{human-scene_act}. Throughout the process, we detected arm poses and employed behavioral prediction algorithm to estimate occluded points. The derived pose estimation enabled the computation of the sleeve states within the arm coordinate system for control. As shown in Fig. \ref{predict-err}, we recorded the progress scalars and sleeve-to-arm distances for both sleeves, and the maximum garment deformation force. Notably, at approximately $15s$, the subject completed donning of the first sleeve and adjusted the stance, inducing observed changes in the sleeve state. Subsequently, our system accurately tracked arm positions and adapted to human motion, successfully completing the secondary sleeve donning and subsequent dressing tasks. These experiments conclusively validate the efficacy of our assisted dressing algorithm in practical applications.

\section{Conclusion}
In this paper, to address the intricate garment-body interaction in robotic-assisted dressing, we propose a novel explicit iterative differentiable cloth simulation. Along with the multi-stage control algorithm, we effectively exploit the contact to adjust human poses with cloth deformation, successfully achieving sequential dressing of different coat sleeves. Regarding cloth simulation, designed high-order bias terms are introduced to enhance computational stability while maintaining iterative precision, thereby accelerating the forward simulation to an $\mathcal{O}(n^2)$ complexity parallel computation. For the dressing control, we augment our differentiable model solution with a constraint-based local cloth model and develop a constrained DDP algorithm to handle the high-dimensional state constraints, thus achieving real-time controller update. Finally, we validate our approach in real-world scenarios, demonstrating its efficacy and practical feasibility.
\par
While demonstrating satisfactory performance in coat dressing tasks, our method currently exhibits limitations in handling cloth self-collisions, and the intentionally introduced high-order bias induce more pronounced energy dissipation, making it less suitable for highly dynamic scenarios. In the future work, it is valuable to expand the model's applicability domain to develop more general assisted-dressing and cloth manipulation algorithms. 

\bibliographystyle{Transactions-Bibliography/IEEEtran}
\bibliography{Transactions-Bibliography/IEEEabrv, Transactions-Bibliography/BIB_xx-TIE-xxxx}
\vspace{-1cm}
\end{document}